\documentclass[11pt]{article}

\usepackage{arxiv}

\usepackage[utf8]{inputenc}
\usepackage[T1]{fontenc}

\usepackage{amsmath,amssymb,amsfonts}
\usepackage{amsthm}

\usepackage{graphicx}
\usepackage{xcolor}
\usepackage{colortbl}
\definecolor{mygray}{gray}{0.88}
\usepackage{booktabs}
\usepackage{multirow}
\usepackage{makecell}   

\usepackage[ruled,linesnumbered,vlined]{algorithm2e}

\usepackage{tikz}
\usetikzlibrary{trees}

\usepackage{subfigure}  

\usepackage[numbers]{natbib}

\numberwithin{equation}{section}
\theoremstyle{plain}
\newtheorem{theorem}{Theorem}[section]

\newtheorem{Prop}[theorem]{Proposition}

\theoremstyle{definition}
\newtheorem{definition}[theorem]{Definition}

\usepackage[colorlinks=true,linkcolor=blue,citecolor=blue,urlcolor=blue]{hyperref}

\title{Frequent subgraph-based persistent homology for graph classification}

\author{Xinyang Chen\\ 
Harbin Institute of Technology, Shenzhen, China \\ 
Universit\'e de Lille, Laboratoire Painlev\'e (UMR CNRS 8524), France \\ 
xinyang.chen.etu@univ-lille.fr \\[2mm]
Ama\"el Broustet \\ 
Universit\'e de Lille, Laboratoire Painlev\'e (UMR CNRS 8524), France \\[2mm] 
Guanyuan Zeng \\ 
Harbin Institute of Technology, Shenzhen, China \\
23S058007@stu.hit.edu.cn\\[2mm]
Cheng He \\ 
Sun Yat-Sen University, Guangzhou, China\\ 
Great Bay University, Dongguan, China \\ 
hech66@mail2.sysu.edu.cn \\[2mm] 
Guoting Chen \\
Harbin Institute of Technology, Shenzhen, China \\ 
guoting.chen@univ-lille.fr 
}

\date{}
\begin{document}
\maketitle






\begin{abstract}
Persistent homology (PH) has recently emerged as a powerful tool for extracting topological features. 
Integrating PH into both machine learning and deep learning models enhances their topology-awareness and interpretability.
However, most PH methods on graphs rely on a limited set of filtrations (e.g., degree- or weight-based), which overlook richer features such as recurring information across the dataset, thereby restricting their expressive power.  
In this work, we propose a novel filtration on graphs, called \emph{\underline{F}requent \underline{S}ubgraph \underline{F}iltration} (FSF), which is derived from frequent subgraphs and produces stable and information-rich \emph{\underline{F}requency-based \underline{P}ersistent \underline{H}omology} (FPH) features.
We explore the theoretical properties of FSF and provide proofs and experimental validation of them.
Beyond persistent homology itself, we further introduce two approaches for graph classification: (i) an FPH-based machine learning model (FPH-ML), and (ii) a hybrid framework integrating FPH with graph neural networks (FPH-GNNs) to enhance topology-aware graph representation learning.
Our proposed frameworks demonstrate the potential for bridging frequent subgraph mining and topological data analysis, providing a new perspective on topology-aware feature extraction and graph representation learning.
Experimental results show that FPH-ML achieves competitive or superior accuracy compared to kernel-based and degree-based filtration methods.
When injected into GNNs, FPH delivers relative gains of ~0.4–21\% (up to +8.2 pts) over their GCN/GIN backbones across benchmarks.
\end{abstract}


\maketitle

\section{Introduction}

Graph classification is fundamental to graph learning, with applications in social networks, cheminformatics, and bioinformatics \citep{liu2023survey}. Existing approaches can be categorized into two main categories: kernel-based and graph neural network (GNN)-based methods \citep{DAktasAF19}. Kernel methods, including random walk \citep{kashima2004kernels}, graphlet \citep{shervashidze2009efficient}, and Weisfeiler--Lehman (WL) kernels \citep{shervashidze2011weisfeiler}, offer strong theoretical foundations but scale poorly to large datasets. In contrast, GNNs have emerged as the dominant paradigm, learning graph representations end-to-end and achieving state-of-the-art performance on diverse benchmarks \citep{ju2025cluster,shi2025two}.
These models primarily focus on local neighborhood aggregation, effectively capturing local structural information. However, despite these advances, GNNs often struggle to capture global, high-order topological features, which are crucial for distinguishing graphs, especially in domains where topological structure plays a central role.

To overcome these limitations, there has been growing interest in integrating topological data analysis (TDA) with machine learning and GNNs, as graphs are inherently topological structures \citep{immonen2023going,pun2018persistent}. 
Among TDA techniques, Persistent Homology (PH) is a powerful method for extracting multi-dimensional topological features, such as connected components, cycles, and voids. 
PH tracks the birth and death of these features across a filtration sequence, resulting in persistence diagrams that provide a compact and informative summary of the graph’s topological structure. 
Unlike GNNs, which primarily focus on local neighbourhood information, PH is capable of capturing global, high-dimensional topological information, offering complementary insights for graph representation and enhancing the interpretability of models. 

A key component of PH is the filtration function, which defines the order in which graph elements (nodes or edges) are added to construct a sequence of nested simplicial complexes. 
Most existing graph filtrations are primarily based on edge weights (e.g., Vietoris-Rips \citep{petri2013topological}, Dowker sink and source filtrations \citep{chowdhury2018functorial}), which limits their applicability to weighted graphs.
Other approaches, such as degree-based filtrations, rely on simple local properties of vertices and often fail to capture meaningful global topological structures \citep{DAktasAF19}.
However, most existing PH-enhanced GNN methods primarily adopt existing filtrations or learn task-driven filtrations, while neglecting the optimization of PH-internal representations themselves \citep{horn2021topological,yan2025enhancing}.
As a result, the expressive potential of PH is not fully exploited.
Therefore, designing more expressive and diverse filtration strategies is essential for enhancing the representational capacity of PH in graph learning tasks.

To address these limitations, we propose a novel graph filtration based on frequent subgraph patterns. 
Our method mines limited-size frequent patterns across the dataset and constructs a filtration for each graph via isomorphic mapping. 
This design incorporates recurring structural information and captures global dataset-level topology. 
We integrate the filtration into both traditional machine learning and GNN pipelines for graph classification. 
Our main contributions are:

(i) We propose \emph{\underline{F}requent \underline{S}ubgraph \underline{F}iltration} (FSF), the first frequency-driven filtration for graphs, which integrates frequent subgraph patterns information into persistent homology (PH) to capture recurring, stable, and dataset-level topological information.

(ii) We explore and prove the theoretical properties of FSF, including the PH dimension (bounded by pattern size), monotonicity, and isomorphism invariance. 

(iii) We introduce two graph classification approaches: 
(a) a \emph{\underline{F}requency-based \underline{P}ersistent \underline{H}omology} (FPH) machine learning model (FPH-ML), and 
(b) a hybrid framework integrating FPH with graph neural networks (FPH-GNN) to enhance topology-aware graph representation learning.

(iv) Extensive experiments on graph classification benchmarks demonstrate that our methods consistently outperform kernel-based baselines, existing PH-based models, and GNNs on most datasets.

The rest of the paper is organized as follows. In section \ref{RelatedWork}, we present work related to this study. Section \ref{Preliminaries} is devoted to basic definitions. In Section \ref{FSbased-filtration}, we present a new framework for frequent sugraph filtration. In Section \ref{FPH-for-graph}, we propose the frequency persistent homology for graph classification. Experimental results are shown in Section \ref{Experiment}. 
Analysis on the limitations and future work is given in Section \ref{Limitations}. Finally, Section \ref{Conclusion} provides the conclusions

\section{Related work}\label{RelatedWork}

\textbf{Frequent subgraph mining.}  
Frequent subgraph mining (FSM) is a fundamental graph mining task that discovers recurring substructures in datasets \citep{rehman2024study}. 
Early methods, such as AGM \citep{inokuchi2000apriori}, follow an Apriori-style candidate generation, but suffer from high costs of repeated subgraph isomorphism tests. 
To overcome this, gSpan \citep{yan2002gSpan} adopts DFS codes with a depth-first extension strategy, while Grami \citep{Elseidy2014grami} encodes subgraph isomorphism as a constraint satisfaction problem for efficient growth and pruning. 

Since not all frequent subgraphs are meaningful, methods like cgSpan \citep{Shaul2021cgSpan} and FCSG \citep{Chen2024closed} focus on concise representations. 
For dynamic graphs, DyFSM \citep{ChenCCG23} uses a fringe set to incrementally update frequent subgraphs as databases evolve, while He et al.\ \citep{HeCCG25, He2024} propose frameworks for mining sequences in dynamic attributed graphs and extracting credible attribute rules. 
FSM has also been extended beyond simple graphs. for example, Aslay et al.\ \citep{AslayNMG18} study hypergraphs, and FreSCo \citep{Preti2022} generalizes FSM to simplicial complexes. 
More recently, learning-based FSM has attracted increasing attention. SPMiner \citep{ying2024} is the first approach to represent subgraphs in a learned embedding space, leveraging order embeddings to preserve the hierarchical relation between subgraphs and their supergraphs. Multi-SPMiner \citep{ZeghinaLBV23} extends this framework to multi-graph settings.

\textbf{Persistent homology and its application.}  
Topological data analysis (TDA) has gained increasing attention, with persistent homology (PH) as its most widely used tool for tracking topological features across scales \citep{FugacciSIF16}. 
PH quantifies structures such as connected components, loops, and voids \citep{edelsbrunner2008persistent}, typically summarized in persistence diagrams that record feature birth and death times. 
To facilitate interpretation, other representations have been developed, including barcodes \citep{wadhwa2018flat}, persistence landscapes \citep{flammer2024spatiotemporal}, and persistence images \citep{adams2017persistence}. 

PH has been applied in diverse domains, including neuroscience \citep{liang2021analysis}, GIScience \citep{corcoran2023topological}, and time-series analysis \citep{ravishanker2021introduction}. 
For example, Corcoran and Jones \citep{CorcoranJ23} apply PH to GIScience, demonstrating robustness, void detection, and downstream analysis through PH-based signatures. 
Beyond direct applications, PH has been integrated into machine learning \citep{pun2018persistent} and increasingly combined with deep learning \citep{pham2025topological}. 
Persistent homology can enhance neural networks by providing higher-order structural information. For example, Zhao et al.\ \citep{zhao2020persistence} proposed a novel network architecture where PH guides graph neural networks (GNNs) by reweighting message passing between graph nodes. 
TOGL~\citep{horn2021topological} introduced a topological graph layer based on persistent homology and demonstrated that topological features can improve the expressive power of GNNs.
Similarly, Rephine \citep{immonen2023going} is a framework that extends PH by designing a new refined filtration, enabling richer topological features to be incorporated into graph representation.

\textbf{Graph classification.}
Graph classification aims to assign labels to entire graphs. Research has evolved from kernel-based similarity measures to embedding methods and, more recently, expressive and scalable GNNs. 
Early work focused on graph kernels combined with classifiers such as SVMs \citep{kriege2020survey}. 
Representative examples include Random Walk Kernels \citep{kashima2004kernels}, which count matching random walks; 
Graphlet Kernels \citep{shervashidze2009efficient}, which compare distributions of small induced subgraphs; 
and Weisfeiler–Lehman (WL) Kernels \citep{shervashidze2011weisfeiler}, which iteratively relabel neighborhoods to capture richer structures. 
These methods perform well on small and medium datasets but suffer from high computational cost and poor scalability. 
To overcome these issues, graph embedding methods learn low-dimensional vector representations without explicit pairwise kernel computation, e.g., Graph2Vec \citep{narayanan2017graph2vec} provides fixed-size embeddings for entire graphs. 

Graph Neural Networks (GNNs) have recently become the dominant approach for graph classification. 
They follow the message-passing paradigm, where nodes aggregate information from neighbors and a readout operation produces graph-level embeddings \citep{gilmer2017neural}. 
Representative models include GCNs \citep{KipfW16}, which apply spectral filtering; GraphSAGE \citep{hamilton2017inductive}, which uses neighborhood sampling for inductive learning; GATs \citep{velivckovic2017graph}, which incorporate self-attention; and GINs \citep{XuHLJ19}, which achieve Weisfeiler–Lehman-level discriminative power through expressive aggregation.

Based on the aforementioned literature review, we note that although significant progress has been made in frequent subgraph mining, persistent homology, and graph classification, their integration remains limited.
Motivated by these, we integrate FSM and PH to capture globally recurring and higher-order topological information, thereby enabling a more informative representation for graph classification.

\section{Preliminaries} \label{Preliminaries}

A vertex-labeled undirected graph is a tuple $ G = (V, E, \ell) $, where $ V $ is a finite set of vertices, $ E \subseteq \{\{u,v\} \mid u,v \in V, u \neq v\} $ is a set of undirected edges, and $ \ell: V \rightarrow L $ is a label function mapping each vertex to a label from a finite label set $ L $. 
Given two vertex-labeled undirected graphs $ G = (V, E, \ell) $ and $ G' = (V', E', \ell') $. If there exists a bijection $ \phi: V \rightarrow V' $ such that:
 $ (u, v) \in E \Leftrightarrow (\phi(u), \phi(v)) \in E' $,
 $ \ell(u) = \ell'(\phi(u)) $ for all $ u \in V $, then $G$ and $G'$ are isomorphic, denoted as $ G \cong G'$. 
 
 We now define MNI-frequent subgraph. 

\begin{definition}[MNI-Frequent Subgraph]\label{MNI_FS}
Let $ \mathcal{D} = (V, E, \ell) $ be a single vertex-labeled undirected graph, and let $ g = (V_g, E_g, \ell_g) $ be a vertex-labeled subgraph. 
An embedding of $g$ in $\mathcal{D}$ is an injective mapping $\phi: V_g \to V$ that preserves adjacency and vertex labels. 
We denote by $\operatorname{Emb}(g,\mathcal{D})$ the set of all such embeddings. 
For each vertex $u \in V_g$, define the distinct node-image set
$\mathrm{Img}(u) := \{\, \phi(u) \mid \phi \in \operatorname{Emb}(g,\mathcal{D}) \,\}.$
The Minimum Node Image (MNI) support of $ g $ in $ \mathcal{D} $ is 
\[
\mathrm{MNI}(g, \mathcal{D}) = \min_{u \in V_g} \bigl| \mathrm{Img}(u) \bigr|.
\]
Then, $ g $ is called \emph{MNI-frequent} if $\text{MNI}(g, \mathcal{D}) \geq \sigma,$
where $ \sigma \in \mathbb{N} $ is a user-defined minimum support threshold.
\end{definition}
\begin{definition}[Simplicial complex]
Let $V = \{ v_0, v_1, \dots, v_k \}$ be a set of $k+1$ affinely independent points.
Then, their convex hull is called a $k$-simplex, where $k$ is the dimension of the simplex:
\[
[v_0, v_1, \dots, v_k] = \{ x \in \mathbb{R}^n \mid x = \sum_{i=0}^{k} \lambda_i v_i,  \lambda_i \geq 0, \quad \sum_{i=0}^{k} \lambda_i = 1\}.
\]
Let $\mathcal{K}$ be the collection of simplices. If $\mathcal{K}$ satisfies the following two conditions, it is called a simplicial complex:

i) Closure Condition: if $\sigma \in \mathcal{K}$ and $\tau$ is any face of $\sigma$, then $\tau \in \mathcal{K}$.
    
ii) Intersection Condition: for all $\sigma, \sigma' \in \mathcal{K}, \sigma \cap \sigma' = \emptyset \text{ or } \sigma \cap \sigma' \in \mathcal{K} .$
\end{definition}

\begin{definition}[$k$-Chain Group]
Given a simplicial complex $ \mathcal{K} $, let $\mathcal{K}_k$ denote the set of all $k$-simplices in $\mathcal{K}$.  
The $k$-chain group is defined as:
\[
C_k(\mathcal{K}) =\displaystyle \Big\{ \sum_{i} a_i \sigma_i \mid a_i \in \mathbb{Z}, \sigma_i \in \mathcal{K}_k \Big\},
\]
which forms a free Abelian group generated by the $k$-dimensional simplices.
\end{definition}

\begin{definition}[Boundary Operator]
The boundary operator $ \partial_p: C_p \to C_{p-1} $ is defined as:
\[
\partial_k(\sigma) = \sum_{i=0}^{k} (-1)^i [v_0, \dots, \hat{v_i}, \dots, v_k],
\]
where 
$ [v_0, \dots, \hat{v_i}, \dots, v_k] $ represents the $(k-1)$-simplex obtained by removing the $ i $-th vertex. 
An important property of this operator is:
$\partial_{k} \circ \partial_{k+1} = 0, \forall k.$
That is, boundaries have no boundary, meaning that each $(k+1)$-dimensional boundary forms a $k$-dimensional closed chain.
\end{definition}

\begin{definition}[Chain Complex]
The boundary operator connects chain groups $ C_k $ into a chain complex, given by:
\[
\cdots \xrightarrow{\partial_{k+2}} C_{k+1} \xrightarrow{\partial_{k+1}} C_k \xrightarrow{\partial_k} C_{k-1} \xrightarrow{\partial_{k-1}} \cdots \xrightarrow{\partial_1} C_0 \xrightarrow{\partial_0} 0.
\]
\end{definition}

\begin{definition}[$k$-Boundary Group]
Given a simplicial complex $ \mathcal{K} $, the $k$-boundary group is defined as:
\[
B_k(\mathcal{K}) = \operatorname{Im}(\partial_{k+1}) = \{ \partial_{k+1}(\sigma) \mid \sigma \in C_{k+1} \}.
\]
\end{definition}

\begin{definition}[$k$-Cycle Group]
Given a simplicial complex $ \mathcal{K} $, the $k$-cycle group is defined as:
\[
Z_k(\mathcal{K}) = \ker(\partial_k) = \{ c \in C_k \mid \partial_k(c) = 0 \}.
\]
\end{definition}

Since $ B_k \subset Z_k $ and $ Z_k \subset C_k $, this implies that $ Z_k $ consists of $k$-cycles (i.e., $k$-closed chains), while $ B_k $ consists of $k$-boundaries (i.e., $k$-closed chains that are also boundaries). The goal of homology is to distinguish nontrivial cycles from boundaries, leading to the definition of the homology group:

\begin{definition}[Homology Group]
The $k$-th homology group of a simplicial complex is defined as:
\[
H_k(\mathcal{K}) = Z_k(\mathcal{K}) / B_k(\mathcal{K}).
\]
Its rank is called the Betti number, denoted as $ \beta_k $, which represents the number of $k$-dimensional nontrivial topological features:
$\beta_k = \dim H_k(\mathcal{K}).$
\end{definition}

\begin{definition}[Filtration Sequence]
Given a simplicial complex $ \mathcal{K} $, a filtration sequence is a nested sequence of sub-simplicial complexes:
\[
\emptyset = \mathcal{K}_{\epsilon_0} \subseteq \mathcal{K}_{\epsilon_1} \subseteq \mathcal{K}_{\epsilon_2} \subseteq \dots \subseteq \mathcal{K}_{\epsilon_n} = \mathcal{K}.
\]
Each $ \mathcal{K}_{\epsilon_i} $ is a subcomplex of $ \mathcal{K} $, and as the scale parameter $ \epsilon $ increases, the simplicial complex grows. That is,
\[
\epsilon_0 \leq \epsilon_1 \leq \dots \leq \epsilon_{n-1} \leq \epsilon_n.
\]
\end{definition}

Thus, we can introduce the definition of persistent homology:

\begin{definition}[Persistent Homology]
Let $\mathcal{K}$ be a simplicial complex with a filtration sequence
$ [\mathcal{K}_{\epsilon_0} \subseteq \mathcal{K}_{\epsilon_1} \subseteq \mathcal{K}_{\epsilon_2} \subseteq \dots \subseteq \mathcal{K}_{\epsilon_n}].$
For each $ \mathcal{K}_{\epsilon_i} \in \mathcal{K} $, the corresponding $k$-th homology group is $ H_k(\mathcal{K}_{\epsilon_i}) $.
The birth of a topological feature $ \alpha $ is defined as its first appearance in $ \mathcal{K}_{\epsilon_b} $, and its death is when it disappears in $ \mathcal{K}_{\epsilon_d}.$ The persistent homology group is then defined as:
\[
H_k^{b,d} = \operatorname{Im} \left( H_k(\mathcal{K}_{\epsilon_b}) \to H_k(\mathcal{K}_{\epsilon_d}) \right).
\]
\end{definition}

\begin{definition}[Persistence Diagram]
Given a filtered simplicial complex $\{K_t\}_{t \in \mathbb{R}}$, the $p$-dimensional persistent homology captures the birth and death times of $p$-dimensional topological features as the filtration parameter $t$ increases. 
The persistence diagram $\mathrm{PD}_p$ is a multiset of points in $\mathbb{R}^2$ of the form $(b, d)$, where $b$ and $d$ denote the birth time and death time, respectively, of a $p$-dimensional feature, with $b < d \leq \infty$. 
Features with infinite death time are often treated as points at infinity or capped at a fixed value for computational purposes. 
\end{definition}

\begin{figure}[ht]
    \centering
    \includegraphics[width=0.8\textwidth]{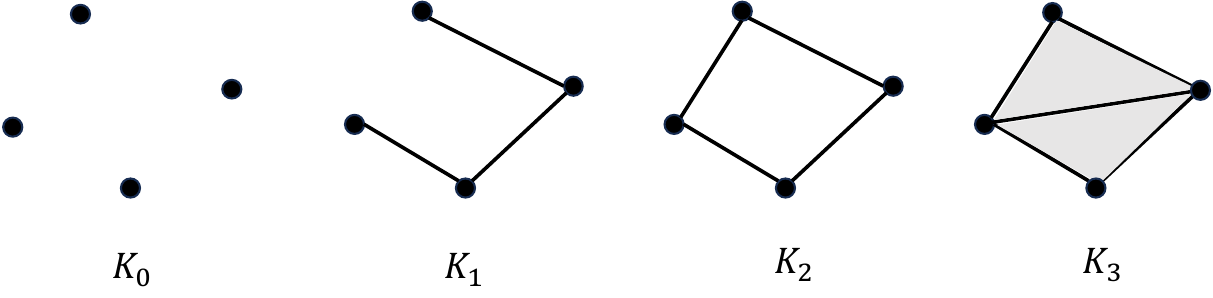} 
    \caption{Example of filtration.}
    \label{fig:examFil} 
\end{figure}

\begin{figure}[ht]
    \centering
    \includegraphics[width=0.5\textwidth]{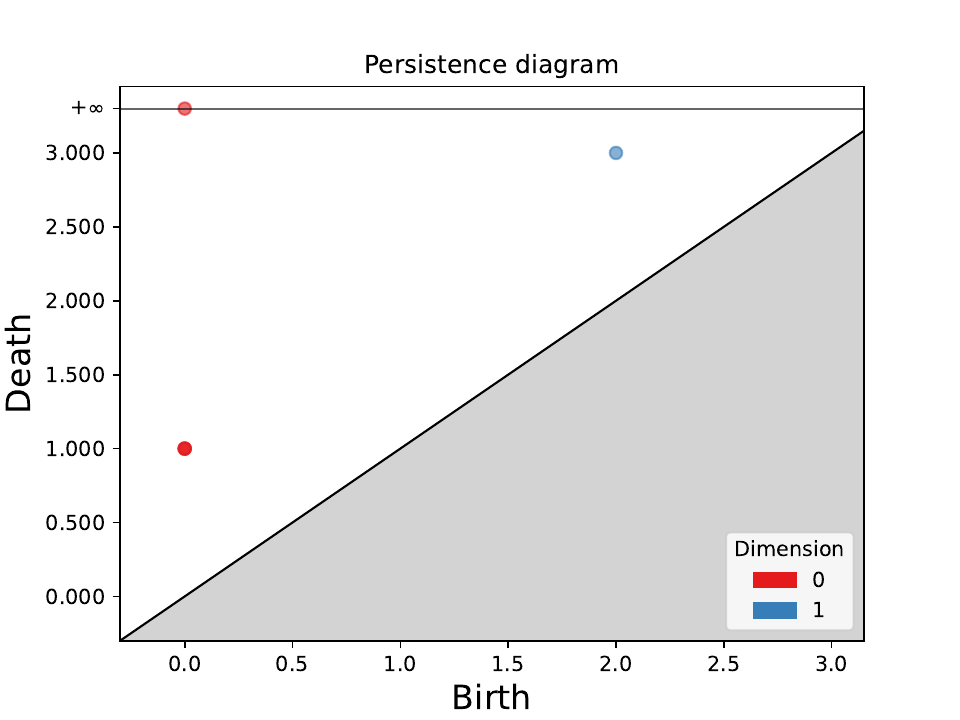} 
    \caption{Example of persistence diagram.}
    \label{fig:examPD} 
\end{figure}

For example, Figure \ref{fig:examFil} illustrates an example of a filtration sequence, and Figure \ref{fig:examPD} shows the persistent diagram. 
For $H_0$ (red points in the persistent diagram), there are four components at filtration 0, but three of them die at filtration 1.
For $H_1$ (blue points in the persistent diagram), a cycle is born at filtration 2, and it dies at filtration 3 because two 2-simplices are added to kill it.

\begin{definition}[Bottleneck Distance]\label{Def:bd}
Let $D_1$ and $D_2$ be two persistence diagrams. 
The bottleneck distance between $D_1$ and $D_2$ is defined as
\[
d_B(D_1, D_2) = \inf_{\gamma} \sup_{x \in D_1} \| x - \gamma(x) \|_{\infty},
\]
where $\gamma$ ranges over all bijections between $D_1$ and $D_2$, and $\| \cdot \|_{\infty}$ denotes the $\ell_{\infty}$ norm. 
Intuitively, the bottleneck distance measures the largest shift needed to match the points of one diagram to the other.
\end{definition}

\section{Frequent subgraph-based filtration}\label{FSbased-filtration} 

\begin{figure}[ht]
    \centering
    \includegraphics[width=0.8\textwidth]{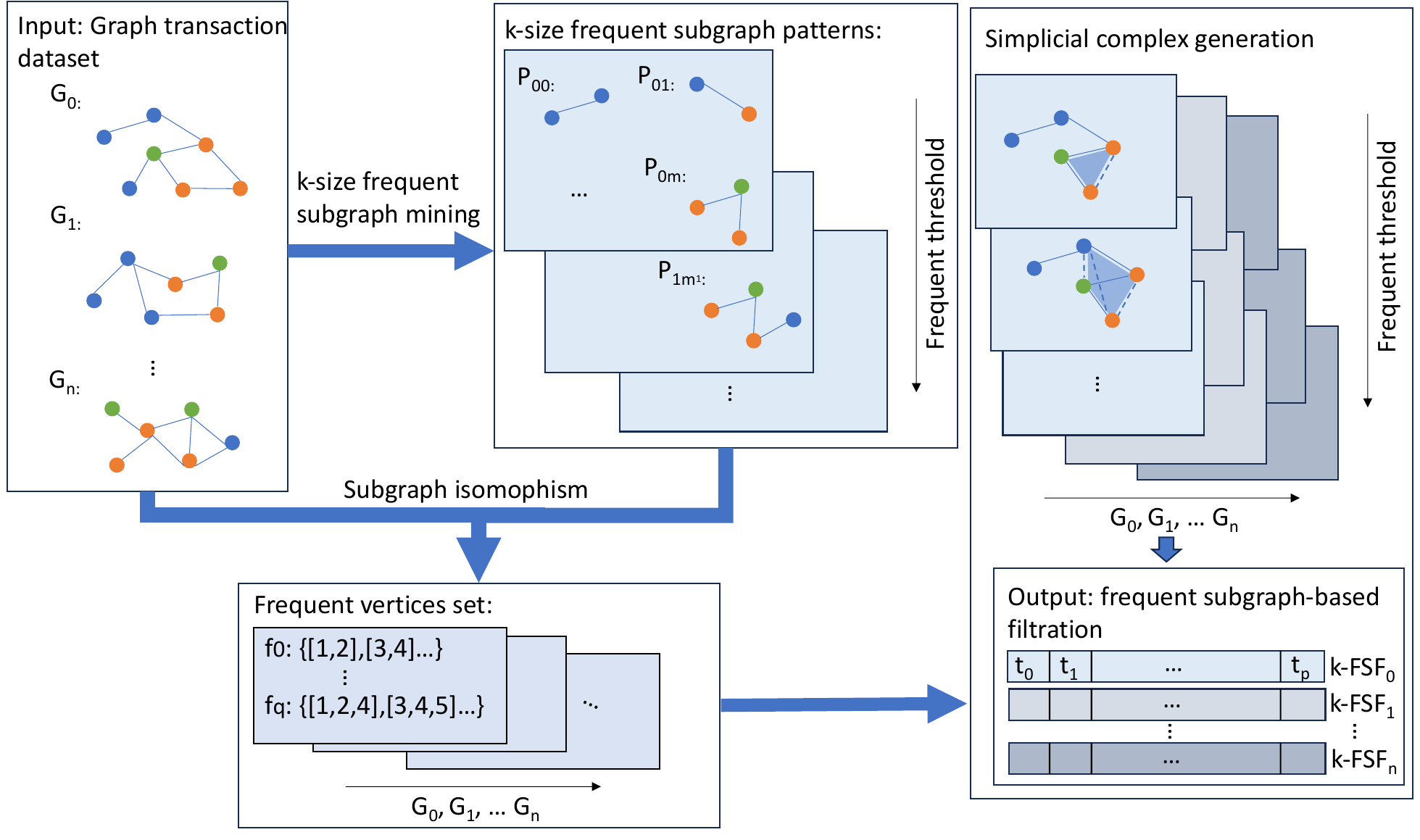} 
    \caption{k-FSF construction.}
    \label{fig:FSFconst} 
\end{figure}

We now propose a novel filtration method constructed from frequent subgraph patterns. 
The framework for filtration construction is shown in Figure \ref{fig:FSFconst}.
Given a vertex-labeled graph transaction dataset $ \mathcal{D} = \{G_0, G_1, \dots, G_n\} $, we consider the union of all graphs as a large graph $ \mathcal{G} $, which serves as the input for FSM.
We now formalize the $k$-size Frequent Subgraph Mining (k-FSM) problem.

Figure~\ref{fig:FSFconst} illustrates the overall process of the filtration construction.

\noindent\textbf{Problem (k-size Frequent Subgraph Mining (k-FSM))} \\
Given a user-defined minimum support threshold $ \sigma \in \mathbb{N} $ and a maximum subgraph size $ k $ (where size refers to the number of vertices in the subgraph), the goal of k-FSM is to discover all subgraphs of size at most $ k $ that are MNI frequent subgraph (Definition \ref{MNI_FS}) in $ \mathcal{G} $. Formally, a subgraph $ g $ is considered k-size frequent if:
\[
\mathcal{F}_k(\mathcal{G}, \sigma) = \{ g \in \mathcal{G}_k \mid \text{MNI}(g, \mathcal{G}) \geq \sigma,\ |V_g| \leq k \}.
\]

We perform k-FSM on $ \mathcal{G} $ to obtain a collection of frequent subgraph patterns, denoted as $ P_{ij} $, where $ i $ indexes the subgraph pattern and $ j $ corresponds to the frequency threshold level. Subsequently, for each graph $ G_i $, we extract the set of vertices that participate in the isomorphisms of frequent subgraph patterns. These vertex sets are then treated as simplices to construct a simplicial complex for $ G_i $. This complex encodes high-order topological relationships among frequently co-occurring vertex groups, thus capturing richer structural information.
Finally, we define a filtration by progressively adding simplices to the complex based on the decreasing order of their corresponding subgraph frequency. This process results in a k-FSF sequence for each graph $ G_i $.

\begin{definition}[k-size Frequent Subgraph-Based Filtration (k-FSF)]
Let $\mathcal{D}$ be a graph transaction dataset and let $G \in \mathcal{D}$ be a graph.
Let $P_t$ denote the set of frequent subgraph patterns whose dataset-level frequency
satisfies $\mathrm{MNI}(p,\mathcal{D}) \ge \frac{1}{t}$ and $|V_p|\le k$.

For a filtration value $t \in \mathbb{R}_+$, we define the simplicial complex $\mathcal{K}_t$ on $G$ as
\[
\mathcal{K}_t=\Big\{\ \triangle\big(\varphi(V_p)\big)\ \Big|\ p \in P_t,\ \varphi \in \mathrm{Emb}(p,G) \Big\}, 
\]
where $\triangle(S)$ denotes the simplex spanned by the vertex set $S$, 
$\mathrm{Emb}(p,G)$ denotes the set of all vertex-label-preserving subgraph
isomorphisms (embeddings) of pattern $p$ into $G$, and $\varphi(V_p)\subseteq V(G)$
is the vertex image set induced by an embedding $\varphi$.

The k-FSF filtration is then defined as the nested family $\{\mathcal{K}_t\}_{t\in\mathbb{R}_+}$.
As $t$ increases (equivalently, as the threshold $1/t$ decreases), more simplices are progressively added to the complex, thereby enlarging the topological structure.
\end{definition}

\begin{figure}[ht!]
    \centering 
    \subfigure[$t_0$]{\includegraphics[width=0.29\textwidth]{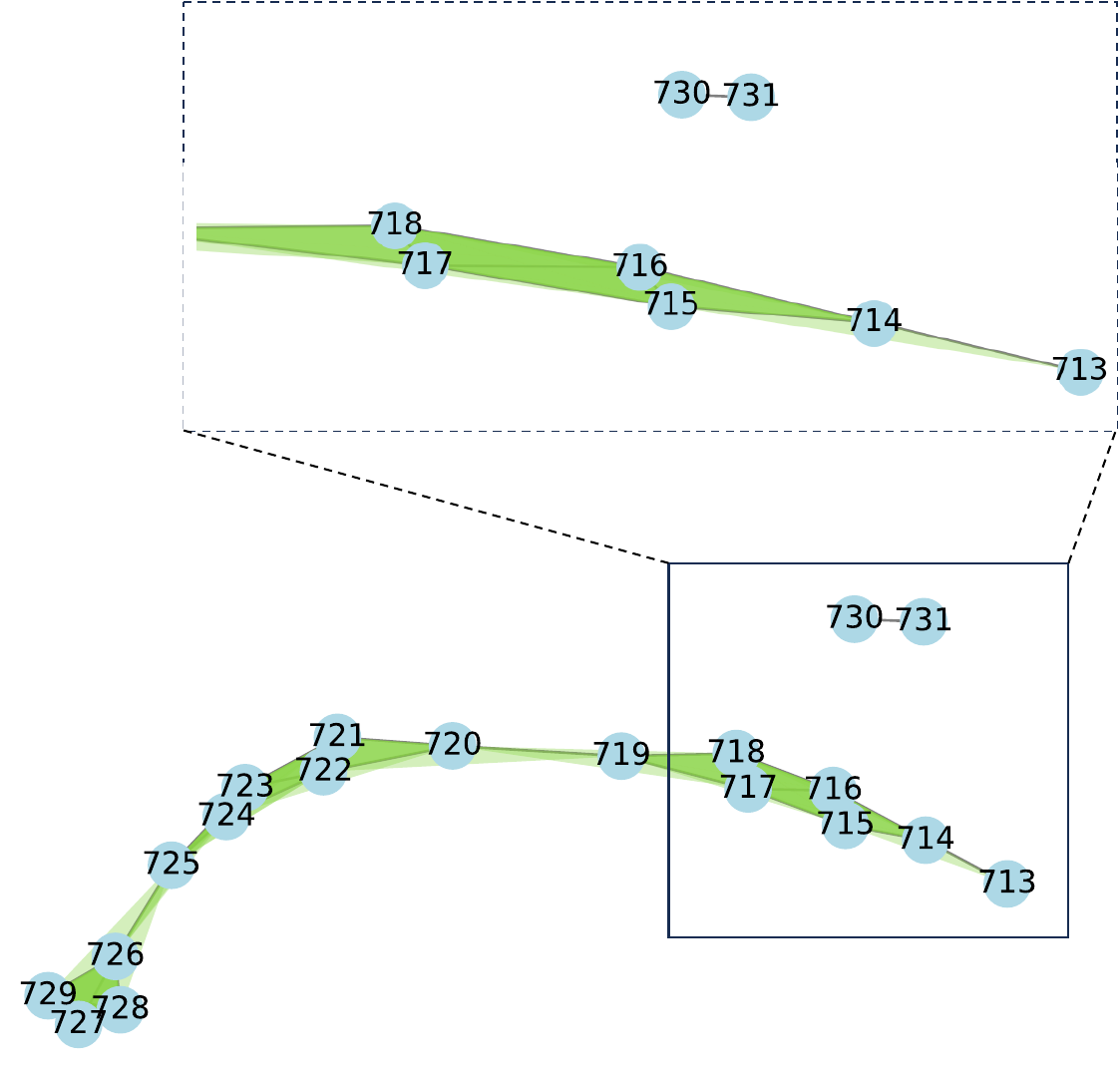} \label{subfig:examFil1}}
    \ \qquad \ 
    \subfigure[$t_1$]{\includegraphics[width=0.29\textwidth]{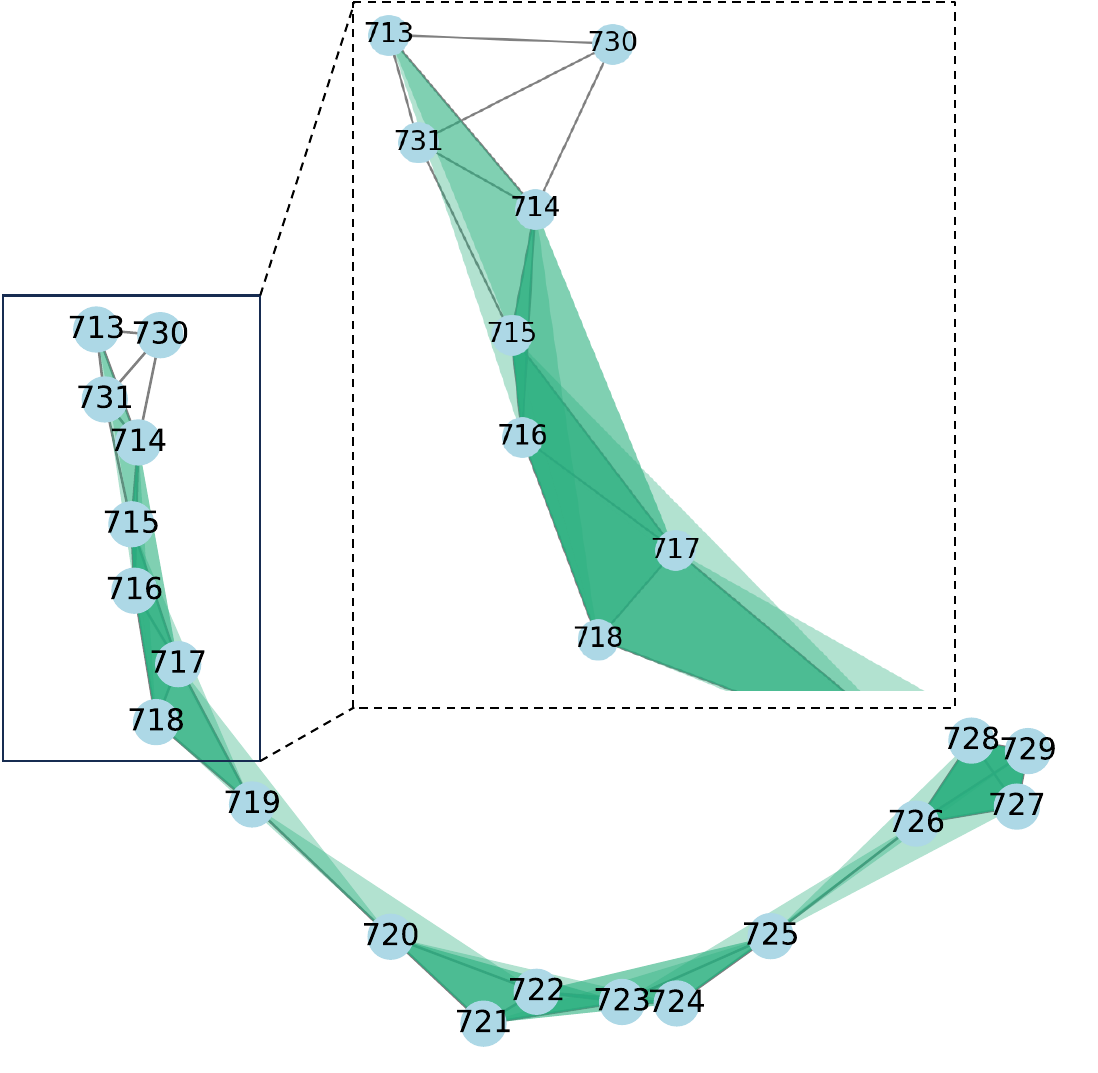} \label{subfig:examFil2}}
    \qquad
    \subfigure[$t_2$]{\includegraphics[width=0.29\textwidth]{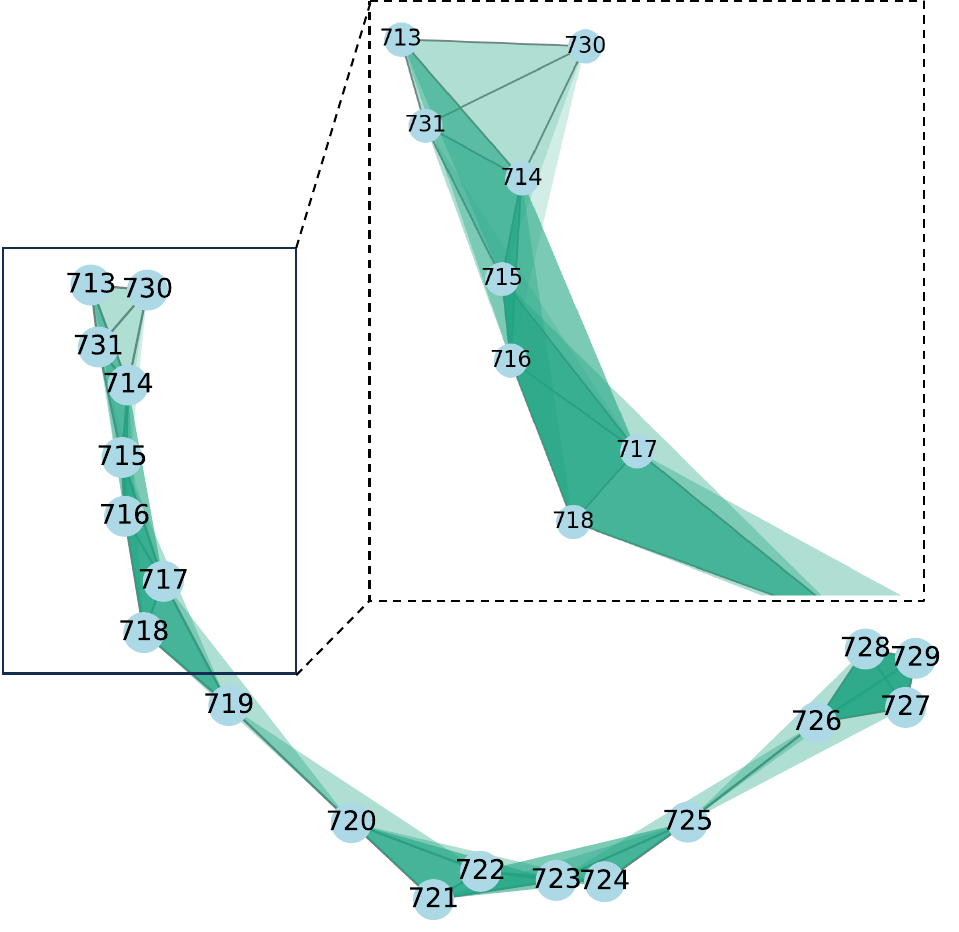}\label{subfig:examFil3}}
    \caption{Example of k-FSF.} \label{fig:examFSF} 
\end{figure}

Figure~\ref{fig:examFSF} illustrates an example of $k$-FSF $\{\mathcal{K}_t\}_{t_0 < t_1 < t_2}$. 
Figures \ref{subfig:examFil1}-\ref{subfig:examFil3} show the simplicial complexes at times $t_0$, $t_1$, and $t_2$, respectively. 
As $t$ increases, additional simplices are incorporated, enriching the topology. 
For example, the isolated component $[730,731]$ at $t_0$ becomes connected to the main complex at $t_1$ through $1$-simplices $[713,730]$, $[714,730]$, and $[713,731]$, as well as a $2$-simplex $[713,714,716]$. 
At $t_2$, further $2$-simplices, such as $[713,730,716]$, are added, increasing topological complexity.

Here, we give three propositions on FSF, and the formal proofs are provided.

\begin{Prop}[Homology Dimension of k-FSF]\label{prop: maxPHdim}
Let $\{\mathcal{K}_t\}_{t \in \mathbb{R}_+}$ be a k-FSF on a graph $G$. Then the maximum dimension $\mathrm{maxD}$ in which persistent homology can be nontrivial is bounded by:
\[
\mathrm{maxD} = \max \{ p \in \mathbb{N} \mid H_p(\mathcal{K}_t) \neq 0 \} \leq k - 1.
\]

Moreover, for $t_1 \leq t_2$, the persistent map $H_{k-1}(\mathcal{K}_{t_1}) \to H_{k-1}(\mathcal{K}_{t_2})$ is 
injective.
\end{Prop}

\noindent\textbf{Proof.}\label{p1}
For every frequent subgraph pattern $p$, if it is isomorphic with a subgraph of $G$, it contributes to vertex sets $T \subset I(V_p)$ of size at most $k$, which induces simplexes $\triangle(T)$ of dimension at most $k - 1$. Thus, for every simplex $\sigma \in \mathcal{K}_t$, the dimension $\dim(\sigma) \leq k - 1.$
Let $C_p(\mathcal{K}_t)$ be the $p$-chain group. Then, we have $C_p(\mathcal{K}_t) = 0$ for all $p \geq k.$
By the definition of $p$-th homology group, 
$H_p(\mathcal{K}_t) = \ker(\partial_p) / \text{Im}(\partial_{p+1}),$
where $\partial_p: C_p \rightarrow C_{p-1}$ is the boundary operator. 
We have
$H_p(\mathcal{K}_t) = 0$ for all $p \geq k.$
Therefore, the maximum dimension for nontrivial persistent homology is $k - 1$.

Furthermore, for $p = k - 1$, the image of the boundary operator $\partial_k$ is 0 (as $C_k = 0$). Hence,
$\text{Im}(\partial_k) = 0.$ Then we have $H_{k-1} = \ker(\partial_{k-1}).$
This implies that any $(k - 1)$-hole, once formed, cannot be filled by higher-dimensional simplices. Therefore, such classes persist indefinitely in the filtration.
\hfill $\square$ 

Note that even for homology groups other than $H_{k-1}$, the groups $H_i(\mathcal{K}_t)$ need not vanish for $t \gg 1$. This contrasts with what happens in filtrations such as Vietoris-Rips. 

\begin{figure}[ht]
    \centering
    \includegraphics[width=0.6\textwidth]{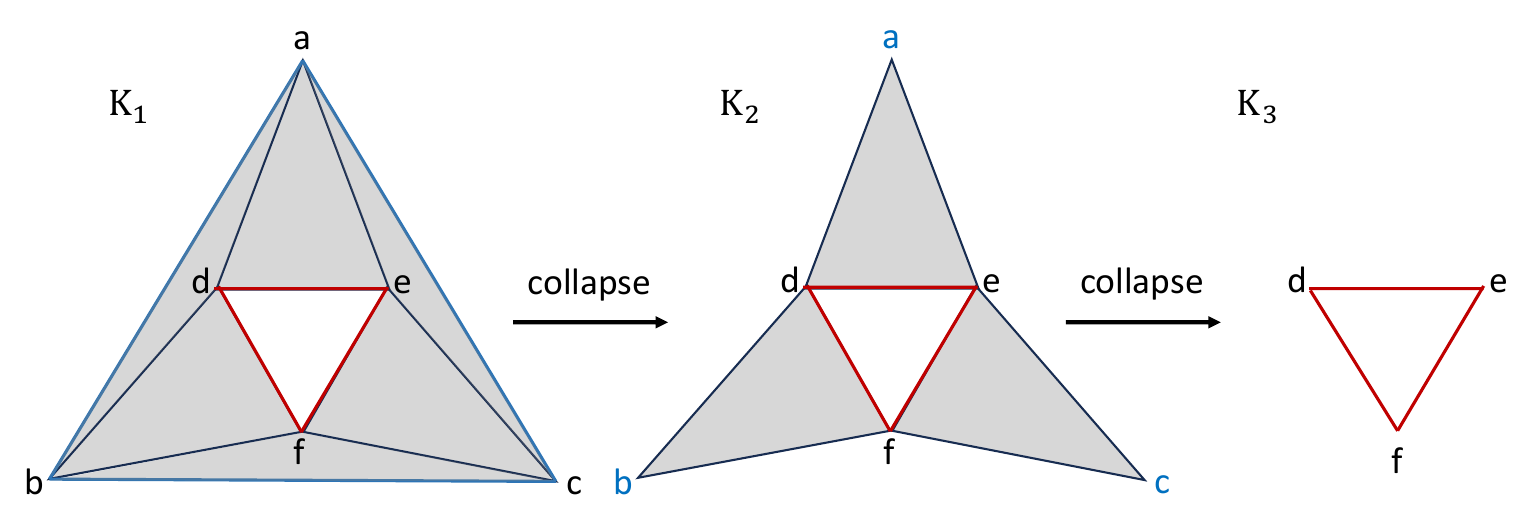} 
    \caption{Example of non-vanishing.}
    \label{fig: nvanishing} 
\end{figure}

\noindent\textbf{Example of non-vanishing.} \label{nv}
As shown in Figure~\ref{fig: nvanishing}, we subdivide the edges 
$\{a,b\}, \{a,c\}, \{b,c\}$ by introducing the vertices $d,e,f$, respectively. 
If the pattern size is equal to $3$, $\forall t \gg 1$, the complex $\mathcal{K}_t$  contains the following $2$-simplices as well as their subsimplices: $\triangle abd$, $\triangle ace$, $\triangle ade$, $\triangle bcf$, $\triangle bdf$, and $\triangle cef$.

Considering $H_1$, in the complex $\mathcal{K}_1$ the $1$-simplex $\{a,b\}$ is only contained in the $2$-simplex $\triangle abc$, hence it is a free face; the same holds for $\{a,c\}$ and $\{b,c\}$. 
By performing elementary collapses on these free faces, we obtain the reduced complex $\mathcal{K}_2$. 

In $\mathcal{K}_2$, the vertices $a,b,c$ become free faces contained uniquely in $\triangle ade$, $\triangle dbf$, and $\triangle efc$, respectively. 
Collapsing them gives the final reduced complex $\mathcal{K}_3$. 

Therefore, we conclude that 
\[
H_1(\mathcal{K}_1) \;\cong\; H_1(\mathcal{K}_3) \;\neq\; 0 .
\]

This shows that persistence obtained from FSF is not only a consequence of the frequency of patterns but also of the topology of the graph.

\begin{Prop}[Monotonicity of FSF]\label{prop:monotonicity}
Let $\{\mathcal{K}_t\}_{t \in \mathbb{R}_+}$ be k-FSF constructed on a graph $G$. Then the filtration is monotonic with respect to $t$, i.e.,
\[
\mathcal{K}_{t_1} \subseteq \mathcal{K}_{t_2} \subseteq \cdots \subseteq \mathcal{K}_{t_n} = \mathcal{K} \quad \text{for } t_1 < t_2 < \cdots < t_n.
\]
\end{Prop}

\noindent\textbf{Proof.}
As $t$ increases, the frequency threshold $\theta = \frac{1}{t}$ decreases, allowing more subgraph patterns to be considered frequent:
\[
t_1 < t_2 \quad \Rightarrow \quad \frac{1}{t_1} > \frac{1}{t_2} \quad \Rightarrow \quad P_{t_1} \subseteq P_{t_2},
\]
where $P_t$ denotes the set of subgraph patterns frequent at threshold $1/t$.

For each frequent subgraph pattern $ p \in P_t $, let $ \mathrm{Emb}(p, G) $ denote the set of all isomorphic embeddings of $ p $ into $ G $. Each such embedding induces a vertex set $ T \subseteq V(G) $, which defines a simplex $ \triangle(T) $ in the simplicial complex.
Thus, as $t$ increases, more subgraph patterns and their embeddings are included, contributing additional simplices to the complex:
\[
\mathcal{K}_{t_1} \subseteq \mathcal{K}_{t_2}, \quad \text{for all } t_1 < t_2.
\]

Hence, the filtration is monotonic in $t$.
\hfill $\square$

\begin{Prop}[FSF is Isomorphism Invariant]\label{prop:iso_inva}
Let $G$ and $G'$ be two isomorphic vertex-labeled graphs, written $G \cong G'$. 
Then the FSFs constructed from $G$ and $G'$ are isomorphic; 
that is, for each filtration step $i$,
\[
\mathcal{K}_i \;\cong\; \mathcal{K}'_i,
\]
where $\mathcal{K}_i$ denotes the simplicial complex at filtration value $i$.  
Consequently, the persistence diagrams of $G$ and $G'$ are equal:
\[
D_G \;=\; D_{G'}.
\]
\end{Prop}

\noindent\textbf{Proof.}
Let $G$ and $G'$ be two graphs such that $G \cong G'$.  
Let $\phi: V(G) \rightarrow V(G')$ be a vertex-label-preserving graph isomorphism.  
For any subgraph $S \subseteq G$, there exists an isomorphic subgraph $\phi(S) \subseteq G'$.  
This subgraph $\phi(S)$ induces the corresponding vertex set $V_{\phi(S)}$.  
Hence, under the same frequent subgraph threshold, the sets of frequent subgraph vertices coincide, i.e.,
$
V_{S \subseteq G} = V_{\phi(S) \subseteq G'}.
$
Therefore, at each step of the filtration, an isomorphic simplicial complex is obtained.  
In particular, the filtration$ \{\mathcal{K'}_t\} _{t \in \mathbb{R}_+}$ on $G'$ satisfies
$
\mathcal{K}_t \;\cong\; \mathcal{K}'_t \quad \text{for all } t.
$
Thus, the persistence diagrams are identical:
$
D_G = D_{G'}.
$
\hfill $\square$

\section{Frequency persistent homology for graph classification}\label{FPH-for-graph}

In this section, we show how FPH features enhance graph classification through two approaches: 
(i) an FPH-based machine learning model, and (ii) integrating FPH with GNNs for enhancing topology-aware representation learning.

\subsection{FPH-based machine learning} \label{sec:nonGNN}
\begin{figure}[ht]
    \centering
    \includegraphics[width=0.98\textwidth]{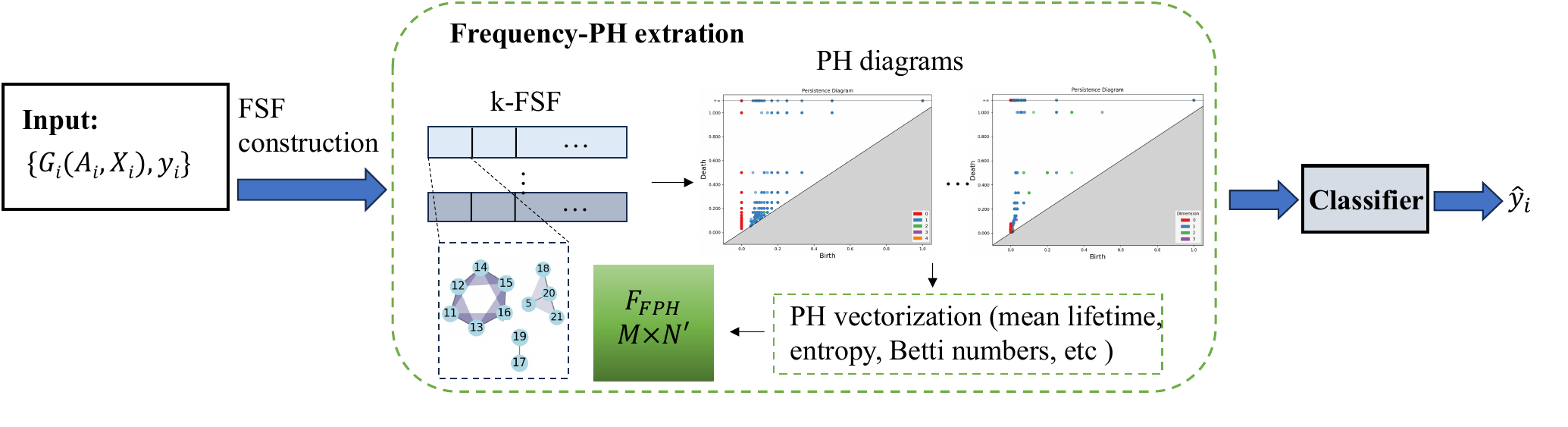} 
    \caption{FPH-ML pipeline.}
    \label{fig:non_gnn} 
\end{figure}

Figure \ref{fig:non_gnn} illustrates the FPH-based machine learning (FPH-ML) pipeline. 
FPH features are extracted via the proposed $k$-FSF, vectorized into PH statistical features, and used as classifier inputs. 
Unlike graph kernels or GNNs, this approach relies solely on topological statistics from persistent homology, without kernels or message passing.

For each homology dimension $k \in \{0,\ldots,d\}$ of graph $G_i$, let 
$\mathrm{PD}_i^{k}=\{(b_j^{k},d_j^{k})\}_{j=1}^{n_k}$ denote the persistence pairs, 
where infinite values are set to $1$. 
The lifetime of each pair is $\ell_j^{k}=d_j^{k}-b_j^{k}$, and if $n_k=0$, the features are zero-padded. 
From the lifetimes, we compute seven statistics: mean ($\mu$), maximum ($L_{\max}$), minimum ($L_{\min}$), median ($\tilde L$), standard deviation ($\sigma$), Betti number ($\beta$), and entropy ($E$).  
The total persistence across all finite pairs is
\(
P_{\mathrm{tot},i}=\sum_{(b,d):\, d<\infty}(d-b).
\)
Concatenating features across all dimensions yields the PH vector
\[
\mathbf{f}_i \;=\; \big[
\mu_i^{0}, L_{\max,i}^{0}, L_{\min,i}^{0}, \tilde L_i^{0}, \sigma_i^{0}, \beta_i^{0}, E_i^{0},
\;\ldots,\;
\mu_i^{d}, L_{\max,i}^{d}, L_{\min,i}^{d}, \tilde L_i^{d}, \sigma_i^{d}, \beta_i^{d}, E_i^{d},
\; P_{\mathrm{tot},i}
\big].
\]
We have $\mathbf{f}_i \in\; \mathbb{R}^{\,7(d+1)+1}.$ Then, all the $f_i$ are stacked as the graph embedding with dimension $M \times N'$:
\[
F_{\text{FPH}} = [f_1; f_2; \ldots; f_M] \in \mathbb{R}^{M \times N'}.
\]

We then train a classifier $h: \mathbb{R}^{N'} \rightarrow \mathcal{Y}$ such that 
\(
\hat{y}_i = h(f_i).
\)

\subsection{Combining FPH with GNNs}
\begin{figure}[ht]
    \centering
    \includegraphics[width=0.95\textwidth]{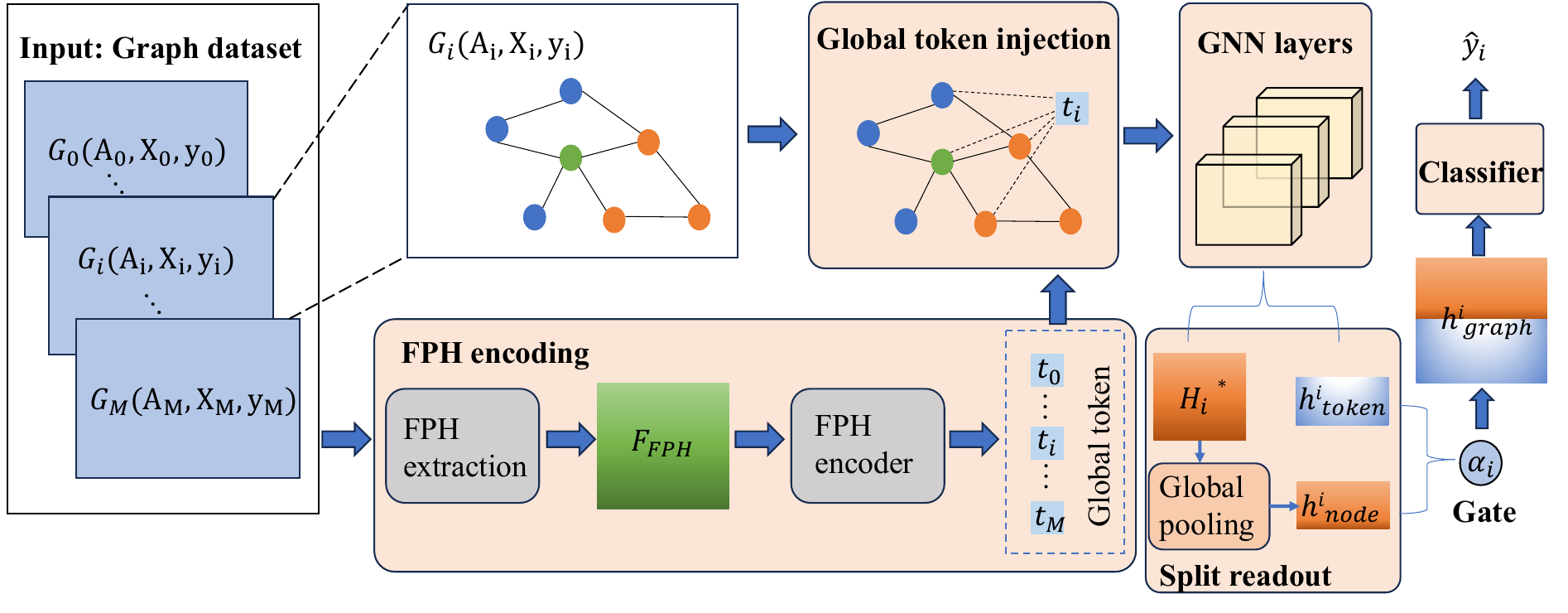} 
    \caption{Overview of the proposed FPH-GNN framework.}
    \label{fig:FPH-GNN} 
\end{figure}

We develop FPH-GNN to enhance the expressiveness of GNN-based graph embeddings by integrating global, stable, and high-order topological information. Figure~\ref{fig:FPH-GNN} illustrates the overall framework.
Specifically, the FPH features are incorporated into the original graph input as a global PH token, guiding the GNN to be aware of global topological structure beyond local message passing. 
 The input consists of a set of graphs  
$\{G_i(A_i, X_i, Y_i)\}_{i=1}^M$,  
where $A_i$ denotes the adjacency matrix, $X_i$ the node feature matrix, and $y_i$ the corresponding graph label. The key components of the framework are detailed below.

\textit{\textbf{FPH encoding.}}  
For each graph $G_i$, we precompute its topological feature $F_{\mathrm{FPH}}^i \in \mathbb{R}^{d_{\text{ph}}}.$ To align these topological features with the latent representation space of node embeddings, we employ a multi-layer perceptron (MLP) as an encoder. Then the FPH embedding is obtained:
\[
e_{\mathrm{fph}}^i = \mathrm{MLP}\!\left(F_{\mathrm{FPH}}^i\right) \in \mathbb{R}^H,
\]
where $H$ denotes the hidden dimension.

\textit{\textbf{Global token injection.}}  
The encoded FPH embedding $e_{\mathrm{fph}}^i$ is injected into $G_i$ as a virtual global token node $t_i$ to serve as a global representative of topological knowledge. 
To maximize its influence, $t_i$ is connected to the top-$K$ highest-degree nodes in $G_i$, forming edges $(t_i, v)$ where $v \in \mathrm{TopKDeg}(G_i)$. 
Consequently, $H_i^{(0)} \in \mathbb{R}^{n_i \times d}$ incorporates both node features and the injected FPH token, enabling subsequent GNN layers to jointly propagate global topological context alongside local structural information.

\textit{\textbf{GNN layers:} }
We use $L$ layers of GNN to propagate both structural and topological information:
\[
H_i^{(l)} = \text{GNN}^{(l)}(A_i', H_i^{(l-1)}) \quad \text{for } l=1,\ldots,L,
\]
where $A_i'$ is the new adjacency including the FPH token, and $H_i^{(l)} \in \mathbb{R}^{n_i \times d}$ denotes the hidden node representations. 

Finally, to guarantee that the encoded FPH features are not overshadowed or diminished after multiple layers of message passing, we use a residual skip connection mechanism from the initial embedding $H_i^{(0)}$ to the final representation $H_i^{(L)}$:
\[
H_i^\ast = \text{ReLU}\!\big(H_i^{(0)} + H_i^{(L)}\big).
\]

\textit{\textbf{Split readout and gated fusion.}}
After $L$ GNN layers, we separately read out the representations of ordinary nodes and the FPH token. 
Since each graph has only one token node, its representation is 
\(
h_{\text{token}}^i = H_{i,t_i}^\ast.
\) 
For the ordinary nodes, we apply a permutation-invariant READOUT function (e.g., global sum pooling):
\[
h_{\text{node}}^i = \text{READOUT}\!\left(\{H_{i,v}^\ast : v \in V_i\}\right).
\]

To adaptively fuse structural and topological information, we compute a gate coefficient $\alpha_i$ via an MLP:
\[
\alpha_i = \sigma\!\left(\text{MLP}_{\text{gate}}\!\left(F_{\text{FPH}}^i\right)\right),
\quad
h_{\text{graph}}^i = h_{\text{node}}^i + \alpha_i \, h_{\text{token}}^i.
\]
The fused representation $h_{\text{graph}}^i$ is then fed to a classifier over $C$ classes, yielding $\hat{y}_i \in \mathbb{R}^C$.

Overall, FPH-GNN combines local structure captured by the GNN with global topological signals from FPH, enabling topology-aware graph representations.

\section{Experiments}\label{Experiment}

We conduct a series of experiments to assess our proposed approach: 
(1) the robustness of FSF,
(2) the performance of FPH-ML, which reveals the discriminative power of FPH features,
and (3) the performance of the proposed FPH-GNN framework.
Here, we set $k=4$, which obtains the maximum PH dimension $H_3$ (Proposition \ref{prop: maxPHdim}). We consider that $H_3$ (once born, never killed) is less informative. Thus, our FPH combines $H_0$, $H_1$, and $H_2$.

The datasets used in our experiments are sourced from the publicly available \texttt{TUDataset} collection \citep{Morris2020TUDataset}, available at {https://chrsmrrs.github.io/datasets/} and the Open Graph Benchmark (OGB)~\citep{hu2020open}.
Table~\ref{tab:datasets} shows the statistics of datasets.
Here, $|G|$ denotes the number of graphs, $|AN|$ the average number of nodes, $|AE|$ the average number of edges, $|VL|$ is the number of node labels, and \emph{Classes} the number of target classes for each dataset.
The implementation is based on PyTorch, and we employ the GUDHI library for persistent homology computations.
To explore the frequent subgraph patterns, we use \textit{gSpan}-based method. All experiments are performed on a CPU workstation equipped with an Intel Core i9 processor and 32 GB of RAM.

\begin{table}[htbp]
\centering
\caption{Dataset statistics.}
\label{tab:datasets}
\begin{tabular}{lccccc}
\toprule
\textbf{Dataset} & $|G|$ & $|AN|$ & $|AE|$ & $|VL|$ & Classes \\
\midrule
AIDS & 2000 & 15.69 & 16.20 & 2 & 2 \\
PROTEINS & 1113 & 39.06 & 72.82 & 3 & 2 \\
NCI1 & 4110 & 29.87 & 32.30 & 37 & 2 \\
ENZYMES & 600 & 32.63 & 62.14 & 3 & 6 \\
DD & 1178 & 284.32 & 715.66 & 89 & 2 \\
ogbg-molhiv & 41127 & 25.5 & 27.4 & 119 & 2 \\
\bottomrule
\end{tabular}
\end{table}

\subsection{Robustness of FSF}\label{sec:stabilityFSF}

To assess FSF robustness, we perturb graphs by randomly adding or removing edges and measure changes via the bottleneck distance between the original and perturbed persistence diagrams. 
Experiments consider $H_1$ and $H_2$ on PROTEINS, NCI1, and AIDS, with perturbations of edge removal (R) or addition (A) at ratios 0.05 and 0.1 of total edges.

\begin{figure}[htbp]
    \centering
     \includegraphics[width=0.29\textwidth]{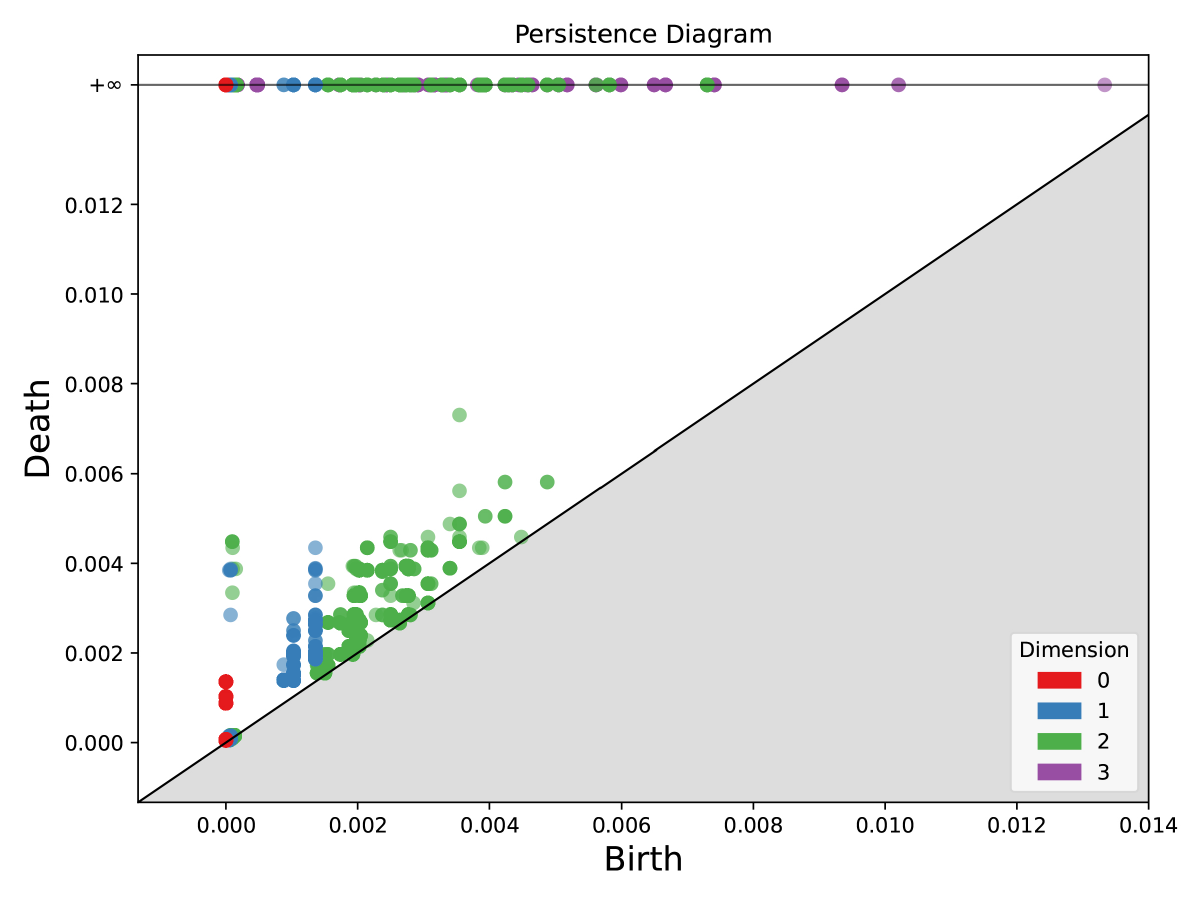}
     \includegraphics[width=0.29\textwidth]{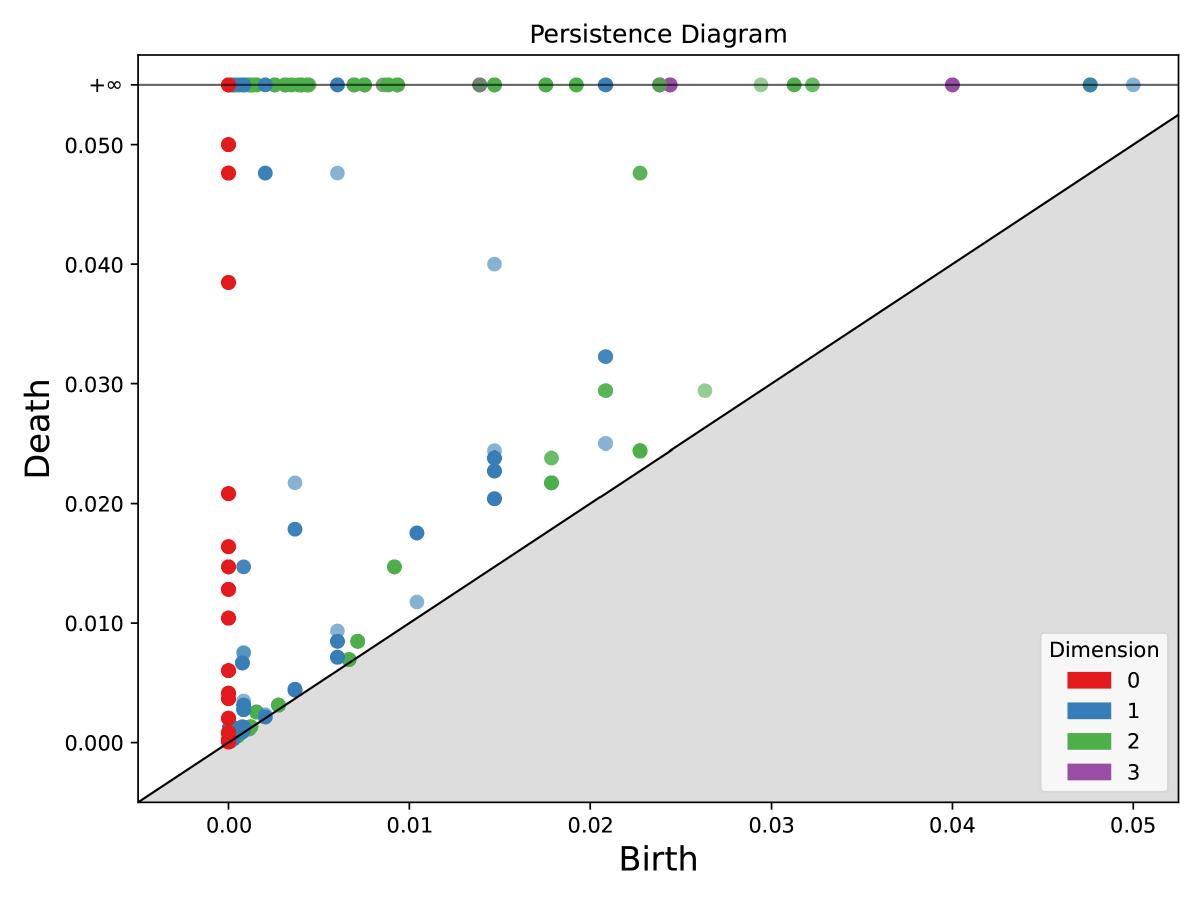}
     \includegraphics[width=0.29\textwidth]{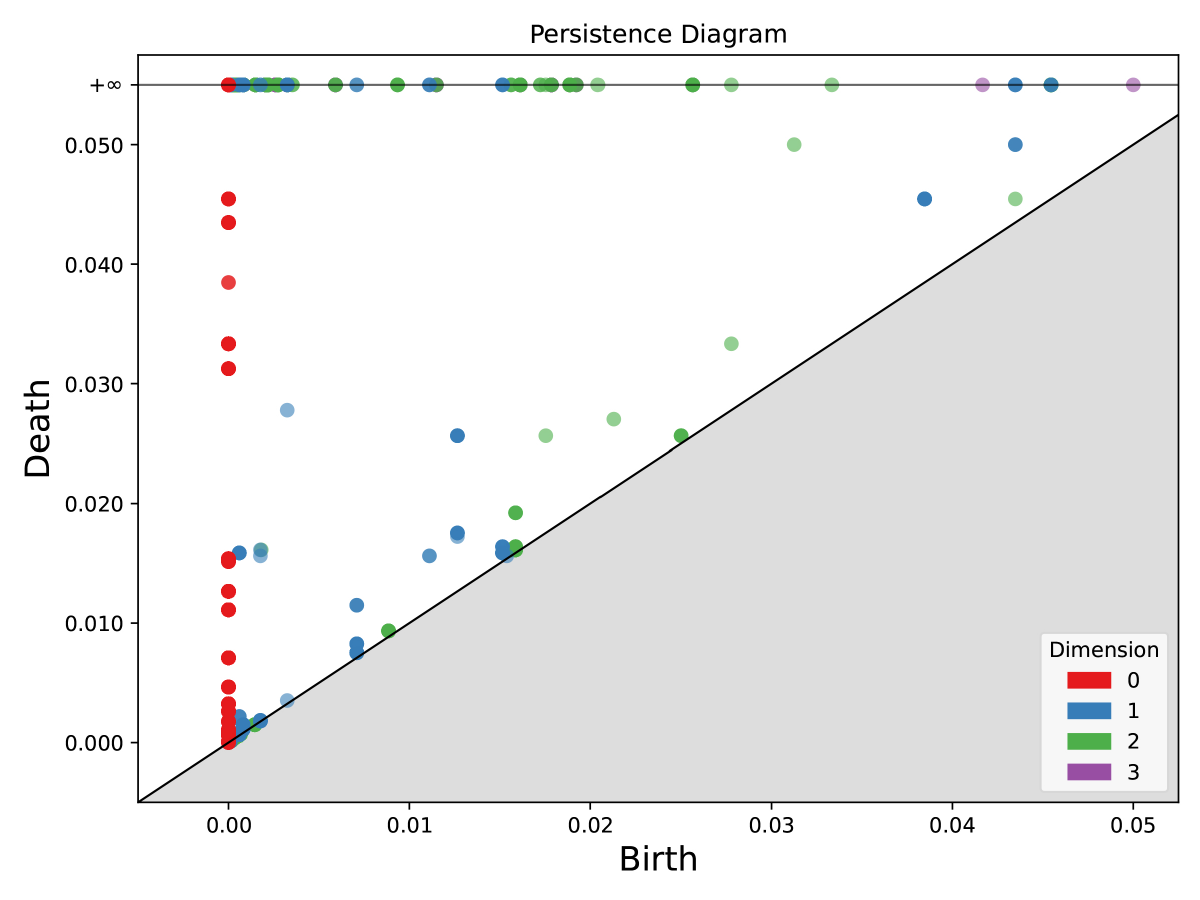}
    \vspace{0.8em}

     \includegraphics[width=0.29\textwidth]{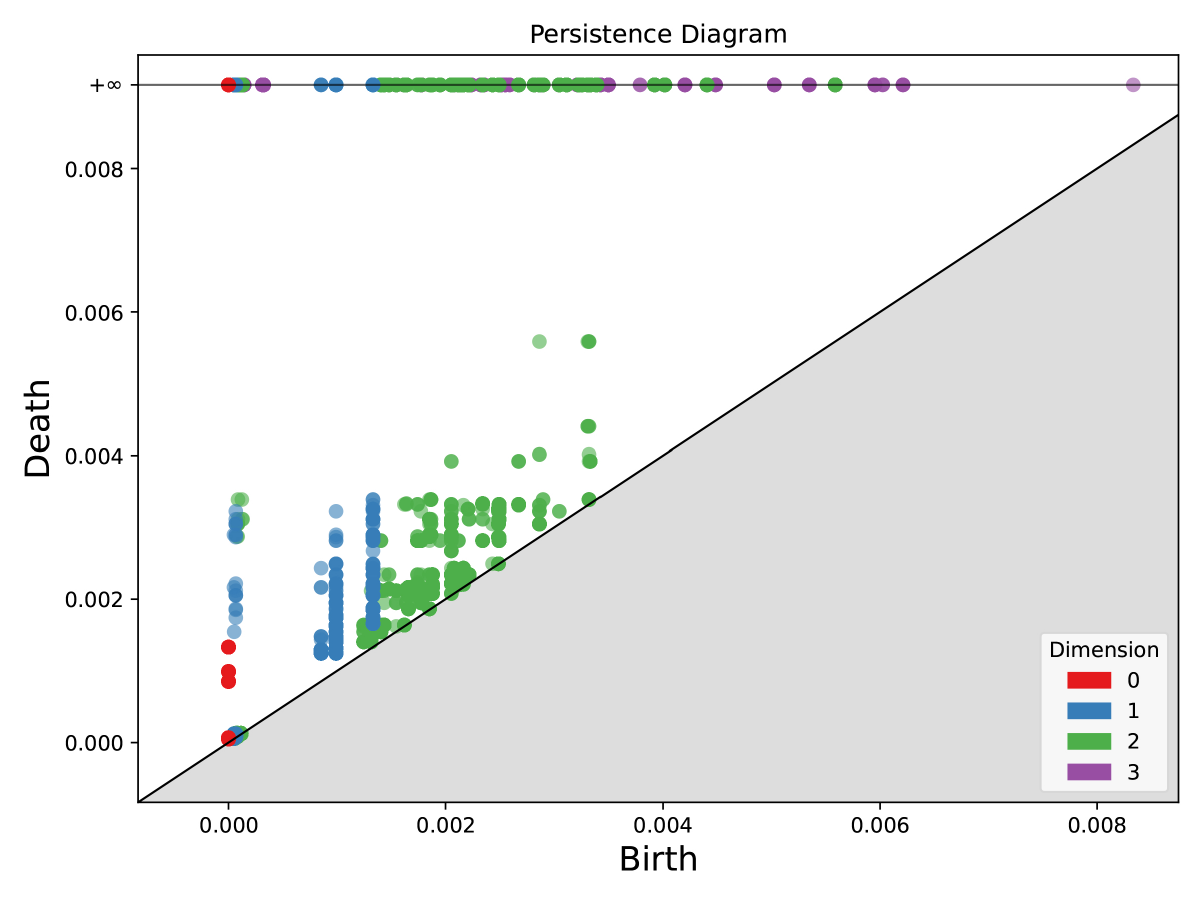}
     \includegraphics[width=0.29\textwidth]{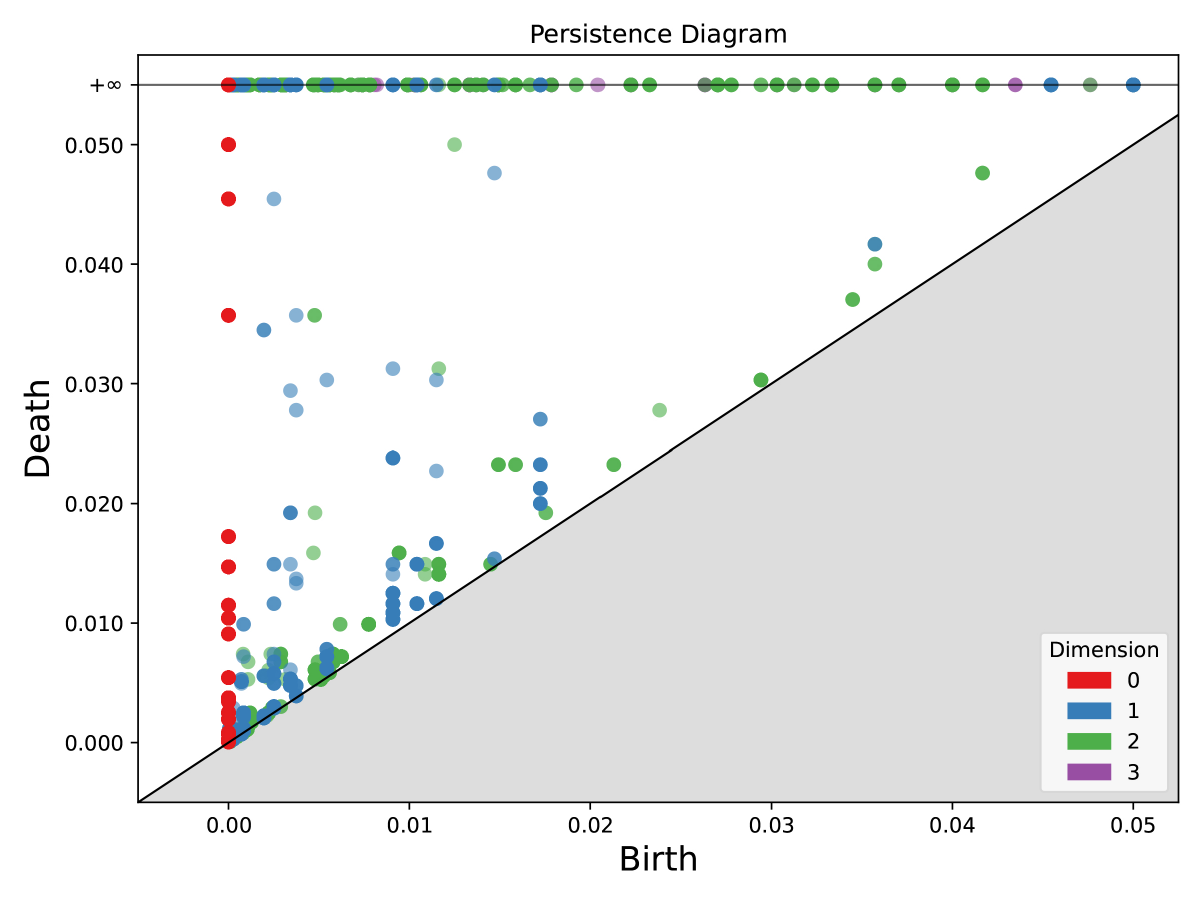}
     \includegraphics[width=0.29\textwidth]{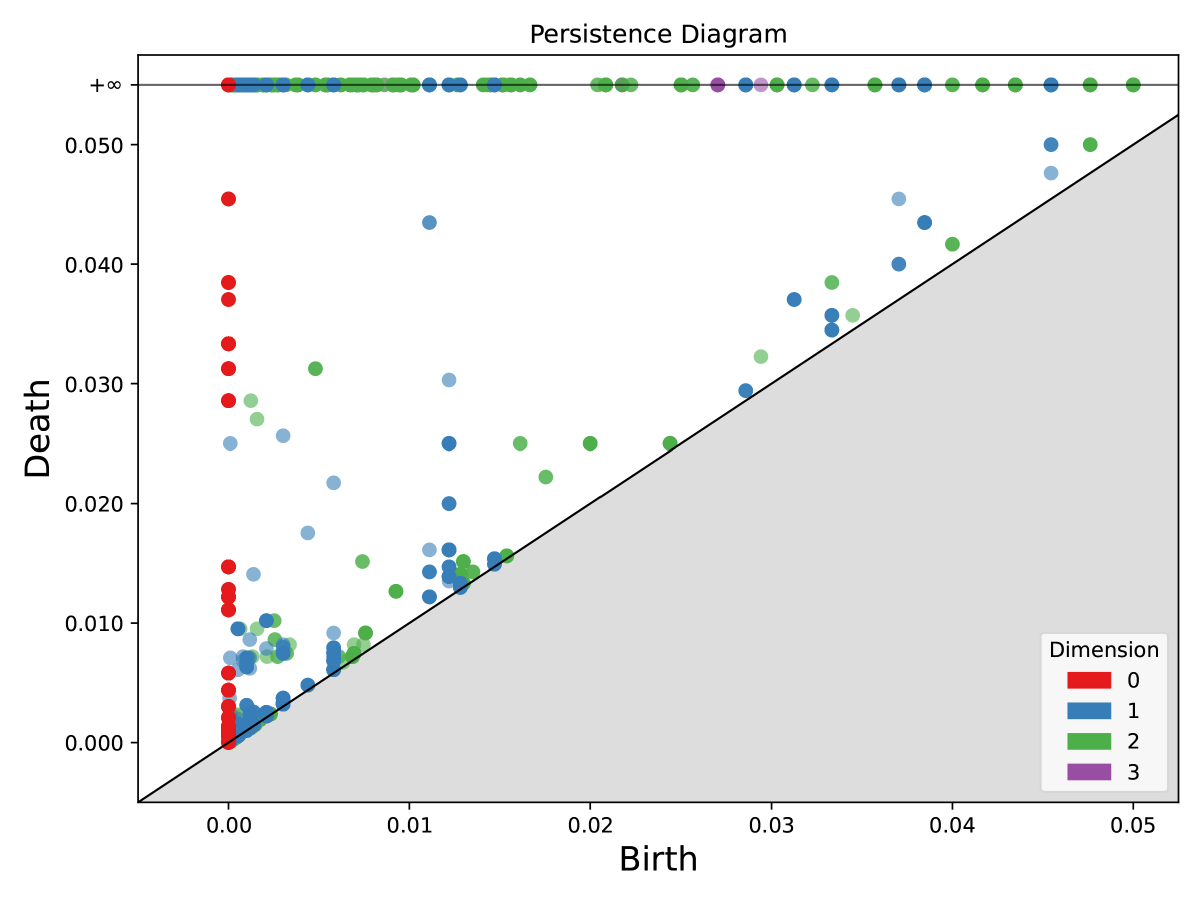}
       \vspace{0.8em}

        \includegraphics[width=0.29\textwidth]{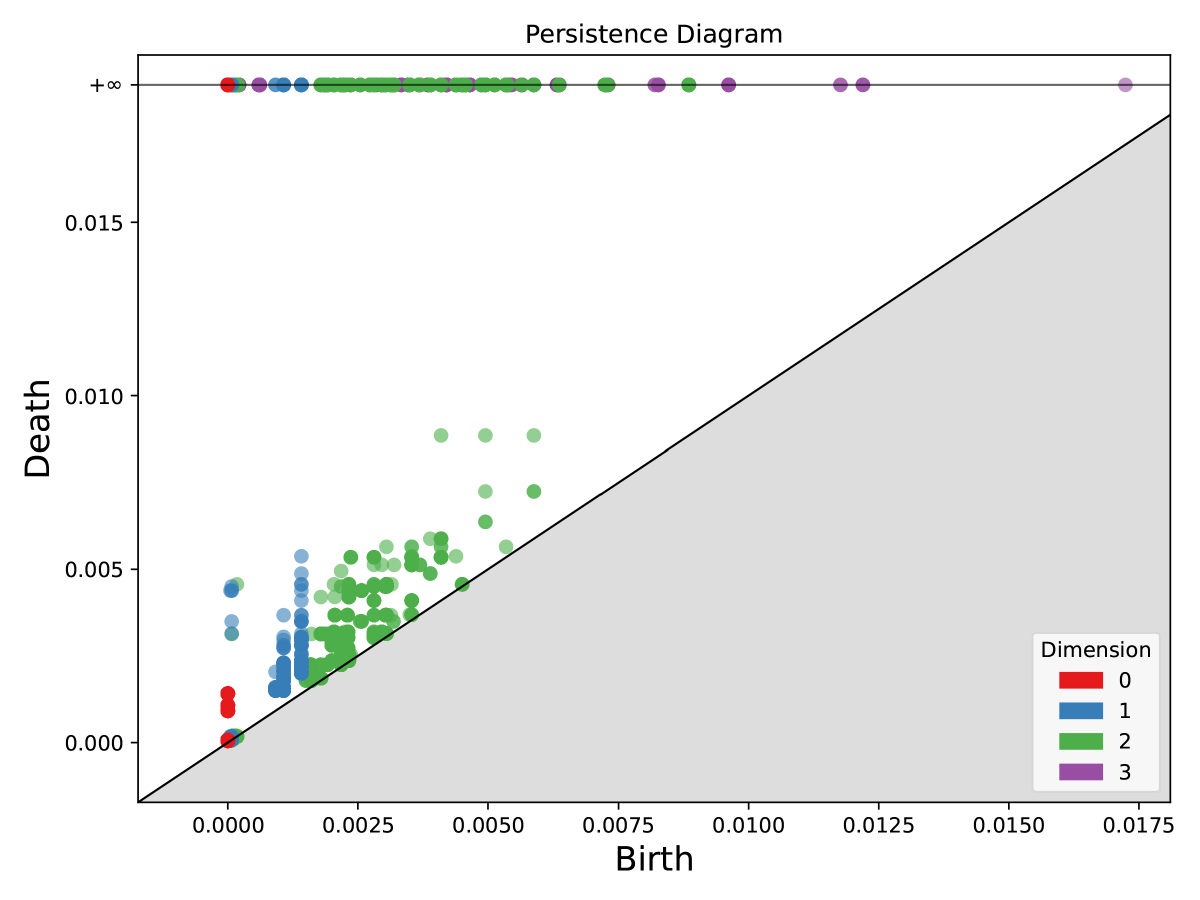} \ 
        \includegraphics[width=0.29\textwidth]{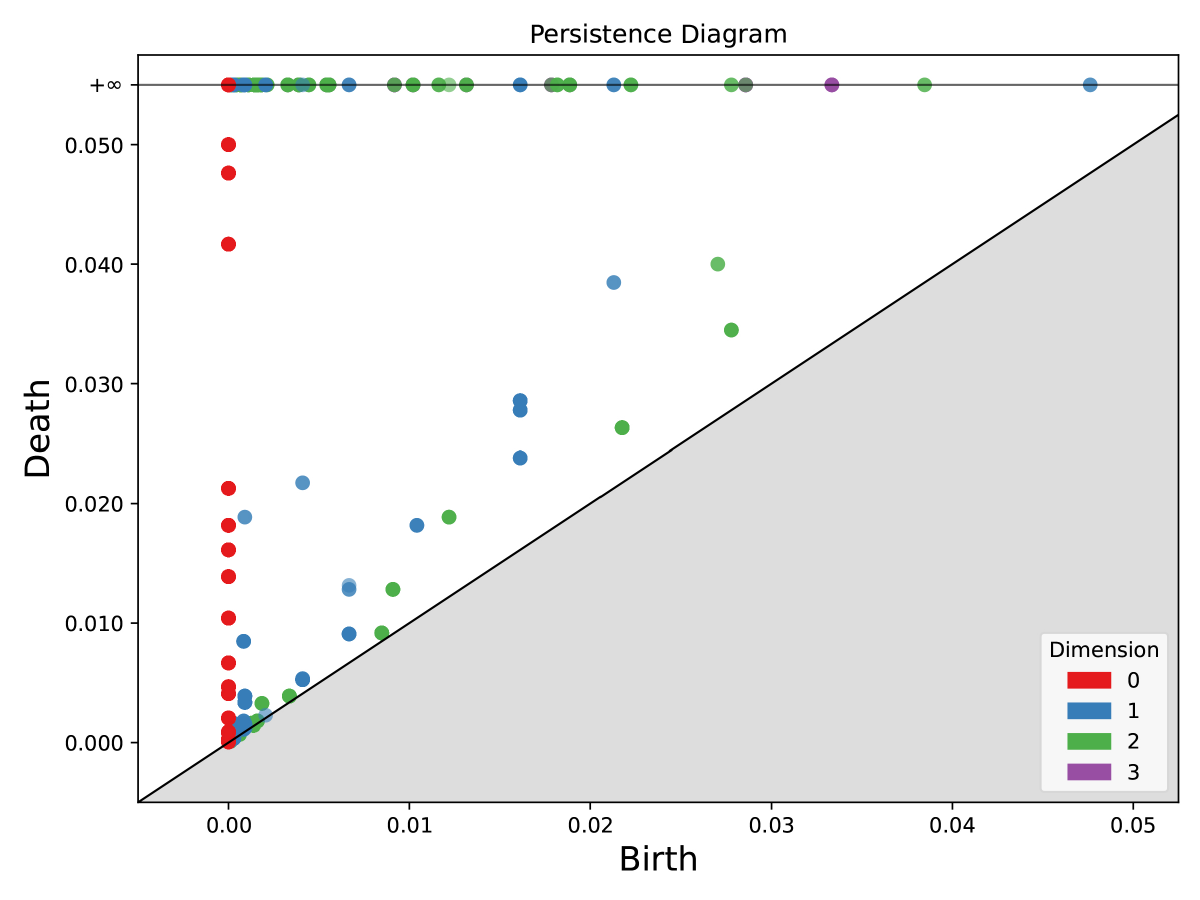} \ 
        \includegraphics[width=0.29\textwidth]{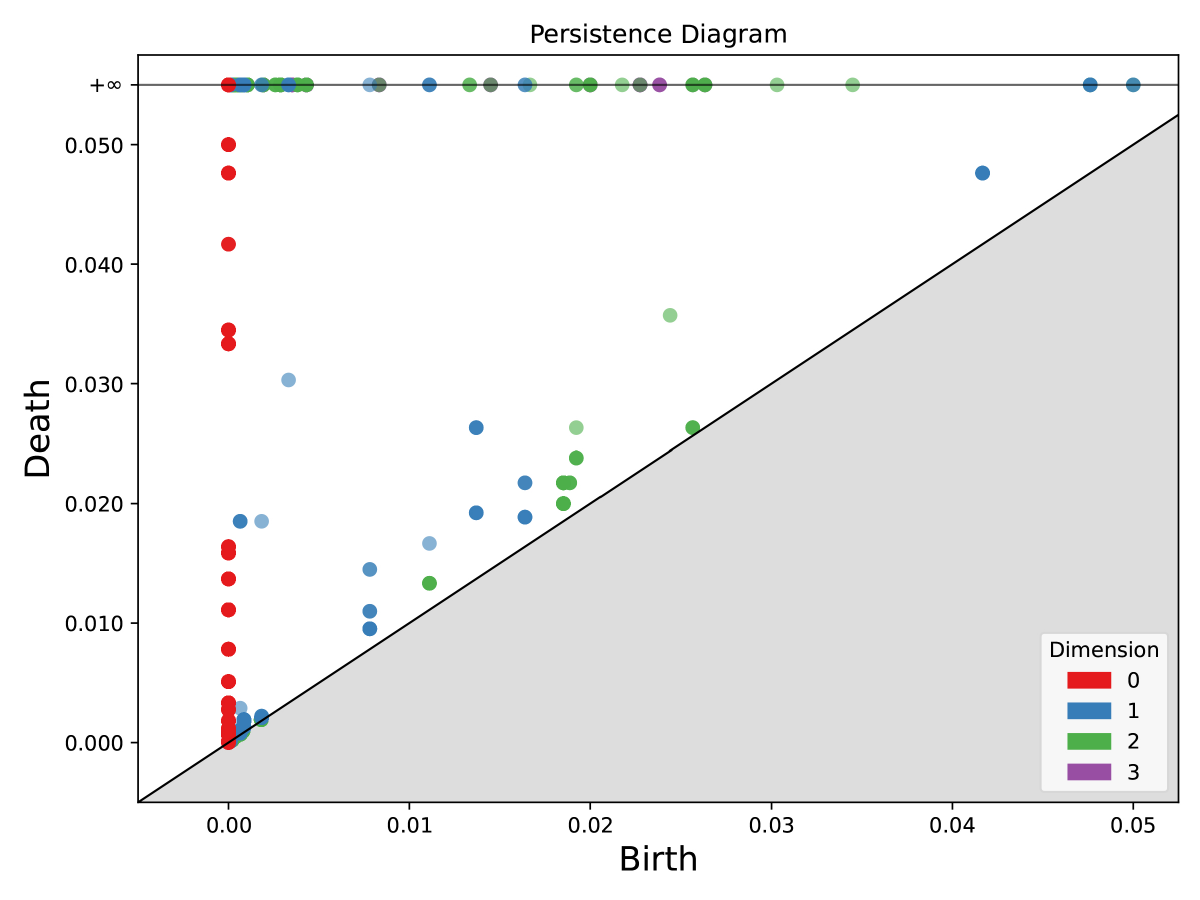}

    \caption{
        Persistence diagrams for three datasets under different perturbations.
        Row 1: Original graphs. 
        Row 2: Graphs after adding 10\% of edges. 
        Row 3: Graphs after removing 10\% of edges.
        Column 1: PROTEINS.
        Column 2: AIDS.
        Column 3: NCI1.
    }
    \label{fig:ph_perturbations}
\end{figure}

Figure \ref{fig:ph_perturbations} shows the persistent diagram of the original graph and the perturbed graphs across these three datasets, and Table  \ref{tab:bt_h1h2} illustrates the bottleneck distance (definition \ref{Def:bd}) results.
As shown in table \ref{tab:bt_h1h2}, the bottleneck distances for all datasets and perturbation settings remain extremely small, typically on the order of $10^{-4}$ to $10^{-2}$. For example, for PROTEINS, the distances for $H_1$ remain below $1.1 \times 10^{-3}$ even with a 0.1 perturbation ratio, and for $H_2$ they remain under $2 \times 10^{-3}$. Across all three datasets, increasing the perturbation ratio from 0.05 to 0.1 leads to only minor increases in bottleneck distance. For instance, in NCI1, $H_1$ exhibits an increase of only about $2 \times 10^{-3}$ for edge removal, while the increase for $H_2$ is about $5 \times 10^{-3}$ . The performance shows a similar trend for edge addition.

\begin{table}[ht]
\centering
\caption{Bottleneck distances for $H_1$ and $H_2$.}
\label{tab:bt_h1h2}
\begin{tabular}{>{\raggedright\arraybackslash}p{1.8cm}c
                p{2.2cm}p{2.2cm}p{2.2cm}p{2.2cm}}
\toprule
\textbf{Dataset} & \textbf{H-dim} & \textbf{R\_0.05} & \textbf{R\_0.1} & \textbf{A\_0.5} & \textbf{A\_0.1} \\
\midrule
PROTEINS & $H_1$ & $9.797 \times 10^{-4}$ & $1.029 \times 10^{-3}$ & $9.629 \times 10^{-4}$ & $9.629 \times 10^{-4}$ \\
         & $H_2$ & $1.604 \times 10^{-3}$ & $1.953 \times 10^{-3}$ & $9.573 \times 10^{-4}$ & $1.719 \times 10^{-3}$ \\
\midrule
NCI1     & $H_1$ & $5.562 \times 10^{-3}$  & $7.221 \times 10^{-3}$ & $1.834 \times 10^{-2}$ & $1.936 \times 10^{-2}$ \\
         & $H_2$ & $4.819 \times 10^{-3}$  & $9.402 \times 10^{-3}$ & $2.230 \times 10^{-2}$ &  $2.276 \times 10^{-2}$ \\
\midrule
AIDS     & $H_1$ & $5.231 \times 10^{-3}$  &  $1.907 \times 10^{-2}$ & $1.435 \times 10^{-2}$ & $1.436 \times 10^{-2}$ \\
         & $H_2$ & $1.245 \times 10^{-2}$  & $1.246 \times 10^{-2}$ & $1.351 \times 10^{-2}$ & $1.355 \times 10^{-2}$ \\
\bottomrule
\end{tabular}
\end{table}

The results demonstrate that the FSF is highly stable under both edge removal and addition, reflecting its robustness to structural noise. 
This robustness arises because the filtration is constructed from frequent subgraph patterns mined from the dataset, which capture stable and recurring topological structures. 

\subsection{Discriminability power of FPH features}\label{sec:disc_fph}
\textbf{FPH-ML performance.} We evaluate the performance of FPH-ML to assess the discriminability power of FPH features using a support vector classifier (SVC). We tried both linear and Radial basis function (RBF) kernel and observed that RBF consistently performed better, thus we use RBF as the kernel.  
We compare FPH against two baselines: 
(i) degree-based PH (DPH) and
(ii) the Weisfeiler–Lehman (WL) kernel. 
Additionally, comparing with random-label experiments (FPH-RL and DPH-RL) further confirms that the performance is not due to chance or label distributions.

For each dataset, we use the same 10-fold cross-validation and  perform an inner 3-fold cross-validation on the training fold to tune the SVM hyperparameters via
GridSearchCV. The best hyperparameters selected on the inner folds are then used to retrain the SVM on the full training fold, and the resulting model is evaluated on the test fold. We report the mean and standard deviation of test accuracy over all 10 folds.

\begin{table}[h]
    \centering
    \caption{Classification accuracy ($\%$) on benchmark datasets using SVC. A \textbf{bold} value indicates the best performance for each dataset.}
    \label{tab:svc_comparison}
    \begin{tabular}{p{1.3cm}ccccc}
        \hline
        \multirow{2}{*}{\textbf{Methods}} & \multicolumn{5}{c}{\textbf{Datasets}} \\
        \cline{2-6}
         & PROTEINS & AIDS & NCI1 & DD & ENZYMES \\
        \hline
        FPH  & \textbf{74.31 $\pm$ 4.78} & \textbf{99.65 $\pm$ 0.59} & 70.00 $\pm$ 1.90 & \textbf{76.65 $\pm$ 3.21} & $36.67 \pm 5.37$ \\
        \makecell[l]{DPH}  & $69.90 \pm 4.89$ & $97.50 \pm 1.20$ & $61.71 \pm 1.59$ & $74.53 \pm 2.84$ & $20.83 \pm 5.39$ \\
        WL & $72.04 \pm 3.36$ & $98.30 \pm 0.75$ & \textbf{74.52 $\pm$ 2.85} & $76.12 \pm 2.63$ & \textbf{51.50 $\pm$ 6.34} \\
        \makecell[l]{FPH-RL} & $50.03 \pm 4.54$ & $51.06 \pm 5.82$ & $46.01 \pm 3.72$ & $43.07 \pm 3.93$ & $17.74 \pm 3.72$ \\
        \makecell[l]{DPH-RL} & $48.05 \pm 3.62$ & $52.01 \pm 4.76$ & $48.08 \pm 5.83$ & $47.02 \pm 3.66$ & $16.03 \pm 4.04$ \\
        \hline
    \end{tabular}
\end{table}

Table~\ref{tab:svc_comparison} presents the results. 
FPH demonstrates superior performance compared to DPH across all datasets, achieving the best accuracy in PROTEINS, AIDS, and DD. 
For instance, on the PROTEINS dataset, FPH reaches 74.31\%, compared to 69.90\% with DPH. 
WL is competitive and achieves the highest performance on NCI1 (74.52\%) and ENZYMES (51.50\%), surpassing FPH. 
The random-label baselines yield drastic performance drops for both FPH and DPH, with accuracies around 16\% for 6-class classification, while 50\% for binary classification, which is close to random guessing. 
These results show that the discriminative power of FPH-ML is not due to chance or label distribution, but the inherent strength of FPH features in graph classification tasks.

\textbf{Contribution of different homology dimensions.} To further assess the power of FPH and the contribution of different homology dimensions, we set $k=5$, which yields $H_3$ features that can potentially be killed.
We evaluate six ablated variants of FPH-ML: (i) combining $H_0$–$H_2$ ($H_{0-2}$), (ii) combining $H_0$–$H_3$ ($H_{0-3}$), and (iii) using each of $H_0$, $H_1$, $H_2$, or $H_3$ individually.

\begin{table}[h]
    \centering
    \caption{Classification accuracy (\%) on benchmark datasets using SVC. 
Gray cells and \textbf{bold} values denote the best performance of combined FPH and individual dimensions, respectively.}
    \label{tab:PHD_dim}
    \begin{tabular}{p{1.3cm}ccccc}
        \hline
        \multirow{2}{*}{\textbf{FPH}} & \multicolumn{5}{c}{\textbf{Datasets}} \\
        \cline{2-6}
         & PROTEINS & AIDS & NCI1 & DD & ENZYMES \\
        \hline
        $H_{0-2}$  & \cellcolor{mygray}74.31 $\pm$ 4.78 & \cellcolor{mygray}99.65 $\pm$ 0.59 & \cellcolor{mygray}70.00 $\pm$ 1.90 & \cellcolor{mygray}76.65 $\pm$ 3.21 & 36.67 $\pm$ 5.37 \\
        $H_{0-3}$  & 73.85 $\pm$ 3.56 & 99.60 $\pm$ 0.62 & 69.83 $\pm$ 2.17 & 76.23 $\pm$ 2.05 &\cellcolor{mygray}39.50 $\pm$ 5.37 \\
        \hline
        $H_0$  & 73.05 $\pm$ 4.36 & \textbf{99.60 $\pm$ 0.70} & \textbf{66.69 $\pm$ 1.37} & 75.04 $\pm$ 2.83 & 27.83 $\pm$ 4.60 \\
        $H_1$ & \textbf{73.23 $\pm$ 3.57} & 90.10 $\pm$ 1.83 & 65.84 $\pm$ 1.92 & \textbf{76.23 $\pm$ 2.17} & 27.33 $\pm$ 6.37 \\
        $H_2$ & 66.13 $\pm$ 4.79 & 80.00 $\pm$ 0.00 & 60.29 $\pm$ 1.81 & 74.79 $\pm$ 2.61 & 29.93 $\pm$ 5.49 \\
        $H_3$ &64.96 $\pm$ 4.63 & 80.25 $\pm$ 1.33 & 55.65 $\pm$ 1.71 & 74.02 $\pm$ 3.35 & \textbf{32.83 $\pm$ 3.80} \\
        \hline
    \end{tabular}
\end{table}
Results in Table~\ref{tab:PHD_dim} show that $H_{0-2}$ slightly outperforms $H_{0-3}$ on most datasets except \textsc{ENZYMES}, and reveal that different datasets rely on different topological characteristics.
For instance, on \textsc{AIDS} and \textsc{NCI1}, $H_0$ dominates, suggesting the importance of connected components, 
while on \textsc{PROTEINS} and \textsc{DD}, $H_1$ achieves the best accuracy, highlighting the role of cycle-related features. 
In contrast, \textsc{ENZYMES} benefits most from $H_3$, corresponding to tetrahedral structures.

\textbf{Discussion.} Since \textsc{ENZYMES} benefits most from higher-order homology, we conducted an additional experiment with $H_4$ and $H_{0-4}$, achieving 29.33\% and \textbf{40.83\%}, respectively. 
This reveals that \textsc{ENZYMES} is closely related to higher-order topological structures. 
However, increasing $k$ will raise the computational cost of both FSM and PH calculation, and $H_{0-2}$ is generally sufficient for most datasets. 
Therefore, we adopt $H_{0-2}$ as the final FPH configuration. 
These results provide a clearer understanding of the role of topological features in graph classification, enhancing the interpretability.

\begin{table}[hb]
    \centering
    \caption{Classification accuracy (\%) on benchmark datasets using GNN-based models. A \textbf{bold} value indicates the best performance for each dataset. A gray background indicates better performance between the model and its corresponding baseline.}
    \label{tab:gnn_comparison}
    \begin{tabular}{lccccc}
        \hline
        \multirow{2}{*}{\textbf{Methods}} & \multicolumn{5}{c}{\textbf{Datasets}} \\
        \cline{2-6}
         & PROTEINS & AIDS & NCI1 & DD & ENZYMES \\
        \hline
        \rowcolor{mygray} 
        FPH-GCN  &  78.61 $\pm$ 3.83 & \textbf{99.65 $\pm$ 0.50} & 78.66 $\pm$ 2.36 & \textbf{82.94 $\pm$ 3.49} & \textbf{47.00 $\pm$ 6.32} \\ 
        DPH-GCN  &  75.81 $\pm$ 4.63 & 99.25 $\pm$ 0.50 & 75.36 $\pm$ 2.96 &76.98 $\pm$ 4.13 &39.81 $\pm$ 8.12 \\ 
        GCN   & 74.48 $\pm$ 1.73 & $99.25 \pm 0.63$ & $74.52 \pm 1.13$ & $75.66 \pm 2.36$  & $38.82 \pm 3.31$ \\
        \hline
        \rowcolor{mygray}FPH-GIN & \textbf{78.80 $\pm$ 3.89} & 99.65 $\pm$ 0.63  & 79.08 $\pm$ 2.45  & 81.92 $\pm$ 2.51  & 44.67 $\pm$ 5.57 \\
        DPH-GIN & 76.35 $\pm$ 2.67 & 99.65 $\pm$ 0.56  & 75.88 $\pm$ 4.21  & 76.52 $\pm$ 3.23  & 42.17 $\pm$ 3.29 \\
        GIN  & $76.16 \pm 2.76$ & 99.25 $\pm$ 0.53 & $75.52 \pm 2.23$   & $76.05 \pm 3.60$  & $42.15 \pm 3.63$ \\
        \hline
        GAT & $74.72 \pm4.01$ & $99.00 \pm 0.75$     & $74.90 \pm 1.83$   & $77.30 \pm 3.68$  & $39.83 \pm 3.68$ \\
        GraphSAGE & $74.01 \pm 4.27$ & $98.20 \pm 1.05$     & $74.73 \pm 1.63$   & $75.78 \pm 3.98$  & $37.93 \pm 3.78$ \\

        Top-$k$Pool  & $75.03 \pm 1.29$ &  $97.25 \pm 0.45$       & $78.92 \pm 2.14$ & $76.95 \pm 2.09$ & $38.35 \pm 3.65$ \\
        
        RePHINE  & $72.32 \pm 1.89$   &  --     & \textbf{80.92 $\pm$ 1.92} & --   & -- \\
        \hline
    \end{tabular}
\end{table}

\subsection{FPH-GNN performance}

In this experiment, we evaluate the performance of the proposed FPH-GNN on five widely used benchmark datasets.
The baselines include classical GNN models: GCN, GIN, GraphSAGE, GAT, a pooling-based method Top-$k$Pool,  a persistent homology-based GNN model RePHINE  \citep{immonen2023going} and the degree-based persistent homology model (DPH-GNN). 
The FPH features are integrated as a plug-in to the GNN models to enhance topology awareness. We perform 10-fold cross-validation, and Table \ref{tab:gnn_comparison}  shows the mean classification accuracy with standard deviation.

The experimental results demonstrate that FPH-GNN models consistently outperform their baseline methods across all benchmarks. For example, FPH-GIN achieves the best performance (78.80\%) on PROTEINS, compared to 76.16\% for baseline GIN.
Furthermore, FPH-GCN attains the best classification accuracy on AIDS (99.65\%), DD (82.94\%), and ENZYMES (47.00\%), surpassing all baselines. 
On the NCI1 dataset, however, RePHINE slightly outperforms our proposed method.  FPH-GCN and FPH-GIN achieve 78.66\% and 79.09\%, respectively, while RePHINE attains the best performance with 80.92\%.

Overall, the experiment results reveal that incorporating FPH features provides complementary global topological information, which can enhance graph-level representation learning, leading to more discriminative performance compared to purely local message-passing approaches.

\subsection{Ablation study of FPH-GNN} \label{A:abla_exp}

To assess the contribution of different components in our proposed FPH-GNN models, we conduct comprehensive ablation studies on PROTEINS, AIDS, NCI1, DD, and ENZYMES, considering both FPH-GCN and FPH-GIN models. 
We evaluate three ablated versions of our model: 
(i) \underline{N}o \underline{G}ate (NG): removing the fusion gate $\alpha$ and directly summing node and token representations; 
(ii) \underline{F}ull \underline{C}onnection (FC): replacing the top-$k$ token injection strategy with full connections from the global topology token to all nodes; 
(iii) \underline{N}o \underline{R}esidual \underline{s}kip \underline{c}onnection (NRsc): eliminating the residual skip connection in the last layer.
\begin{table}[h]
    \centering
    \caption{Ablation study for FPH-GNN.}
    \label{tab:abla_exp}
    \begin{tabular}{p{1.8cm}ccccc}
        \hline
        \multirow{2}{*}{\textbf{Methods}} & \multicolumn{5}{c}{\textbf{Datasets}} \\
        \cline{2-6}
         & PROTEINS & AIDS & NCI1 & DD & ENZYMES \\
        \hline
        FPH-GCN  &  \textbf{78.61 $\pm$ 3.83} & \textbf{99.65 $\pm$ 0.50} &  \textbf{78.66 $\pm$ 2.36} & \textbf{82.94 $\pm$ 3.49} & \textbf{47.00 $\pm$ 6.32} \\ 
        FPH-GCN-NG  &78.08 $\pm$ 4.65&99.60 $\pm$ 0.62&77.92 $\pm$ 4.62&81.36 $\pm$ 2.42&45.67 $\pm$ 5.88 \\
        FPH-GCN-FC &78.52 $\pm$ 3.95&99.60 $\pm$ 0.62 &77.79 $\pm$ 2.55&81.49 $\pm$ 2.99& 42.33 $\pm$ 5.73\\
      FPH-GCN-NRsc &78.35 $\pm$ 3.31&99.60 $\pm$ 0.62&78.64 $\pm$1.70 &82.30 $\pm$ 3.04 &45.17 $\pm$ 4.91 \\
       GCN   & 74.48 $\pm$ 1.73 & 99.25 $\pm$ 0.63 & 74.52 $\pm$ 1.13 & 75.66 $\pm$ 2.36  & 38.82 $\pm$ 3.31 \\ 
        \hline
        FPH-GIN & \textbf{78.80 $\pm$ 3.89} & \textbf{99.65 $\pm$ 0.63}  & \textbf{79.08 $\pm$ 2.45}  & \textbf{81.92 $\pm$ 2.51}  & \textbf{44.67 $\pm$ 5.57} \\
        FPH-GIN-NG  &76.98 $\pm$ 4.58&99.55 $\pm$ 0.60&78.25 $\pm$ 2.80& 80.90 $\pm$ 2.98&43.50 $\pm$ 7.32\\
        FPH-GIN-FC &78.38 $\pm$ 4.78 &99.55 $\pm$ 0.61 &78.22 $\pm$ 1.16&77.93 $\pm$ 1.96&40.67 $\pm$ 7.23 \\
      FPH-GIN-NRsc &77.98 $\pm$ 5.35 &99.55 $\pm$ 0.65 & 78.35 $\pm$ 2.25 & 79.23 $\pm$ 2.98 & 43.25 $\pm$ 5.58\\
        GIN  & 76.16 $\pm$ 2.76 & 99.25 $\pm$ 0.53 & 75.52 $\pm$ 2.23   & 76.05 $\pm$ 3.60  & 42.15 $\pm$ 3.63 \\
        \hline
    \end{tabular}
\end{table}

As shown in Table~\ref{tab:abla_exp}, removing any of these modules consistently reduces performance compared to the full FPH-GNN. 
These results highlight the importance of each design choice. 
For instance, directly summing graph and token embeddings (NG) leads to performance drops, since it removes the adaptive fusion of global and local features. 
Similarly, fully connecting the global topology token (FC) may introduce noise, thereby reducing the informative role of FPH features. 
The residual connection (NRsc) also proves essential, as it enhances the influence of FPH features after several GNN layers. 

A noteworthy observation is that ablating these modules degrades performance less severely than removing the FPH features themselves, 
demonstrating that FPH features are the key factor driving the improvement of FPH-GNN over plain GNN baselines.

\textbf{Frequency features only.}
To further assess the contribution of PH, we conduct an additional feature-replacement study in which FPH features are replaced by frequency features only.
Specifically, we select the top-20 frequent subgraphs identified in the previous FSM step and recompute their MNI frequency in each graph.
This results in a 20-dimensional MNI-based feature vector for each graph.

As shown in Table~\ref{tab:mni_frequency_only}, using frequency features of the top-20 frequent subgraphs already yields competitive performance across all
three datasets, indicating that frequent subgraph statistics capture meaningful
discriminative information.
Nevertheless, FPH-GIN consistently outperforms the frequency-only methods.
This indicates that FPH offers complementary information beyond frequency counts by encoding
richer topological and structural relationships through persistent homology.

\begin{table}[htpb]
\centering
\caption{Replacing FPH with MNI-based frequency features.}
\label{tab:mni_frequency_only}
\begin{tabular}{lccc}
\toprule
\textbf{Methods} & \textbf{PROTEINS} & \textbf{ENZYMES} & \textbf{NCI1} \\
\midrule
Top-$20$ FS-GIN
& $77.36 \pm 3.84$ 
& $42.17 \pm 9.16$ 
& $78.66 \pm 1.57$ \\
FPH-GIN
& \textbf{78.80 $\pm$ 3.89}
& \textbf{44.67 $\pm$ 5.57} 
& \textbf{79.08 $\pm$ 2.45} \\
\bottomrule
\end{tabular}
\end{table}

\subsection{Runtime and scalibility}

In this section, we evaluate the runtime and discuss the scalability of the proposed method. We first compare the component-wise and overall runtime of FPH-ML, FPH-GNN, DPH-ML, WL, and GCN across three datasets of different scales; among them, AIDS is the smallest, PROTEINS represents a medium scale, and DD is the largest. In this comparison, we set the maximum pattern size $k$ of the FSM to 4, which allows the construction of nontrivial three-dimensional persistent homology. 
It is important to note that the pre-computation stage (FSM and PH calculation) of PH-based methods needs to be performed only a single time.

\begin{table*}[htpb]
\centering
\caption{Runtime comparison (in seconds) on three datasets.
FSM and PH are executed once per training split as preprocessing.}
\label{tab:runtime}
\begin{tabular}{lccccc}
\toprule
\textbf{Dataset} & \textbf{Method} & \textbf{FSM} & \textbf{PH} &
\textbf{Train} & \textbf{Total} \\
\midrule

\multirow{4}{*}{AIDS}
 & FPH-ML      &  2.90 & 1.27  & 0.01  & 4.18   \\
 & FPH-GNN     &  2.90 & 1.27  & 36.41 & 40.58  \\
 & DPH-ML      &   --  & 0.39  & 0.02  & 0.41   \\
 & GCN         &   --  &  --   & 34.41 & 34.41  \\
\midrule

\multirow{4}{*}{PROTEINS}
 & FPH-ML      &  26.78 & 6.52 & 0.02  & 33.32  \\
 & FPH-GNN     &  26.78 & 6.52 & 39.41 & 72.71  \\
 & DPH-ML      &   --   & 2.04 & 0.03  & 2.07   \\
 & GCN         &   --   &  --  & 36.41 & 36.41  \\
\midrule

\multirow{4}{*}{DD}
 & FPH-ML      & 738.27 & 543.38 & 0.02   & 1281.67 \\
 & FPH-GNN     & 738.27 & 543.38 & 181.41 & 1463.06 \\
 & DPH-ML      &   --   & 6.05   & 0.05   & 6.10    \\
 & GCN         &   --   &  --    & 178.41 & 178.41  \\
\bottomrule
\end{tabular}
\end{table*}

As shown in Table~\ref{tab:runtime}, both FSM and FPH introduce noticeable
computational overhead; however, the total cost remains manageable for all
datasets evaluated. 
The persistent homology computation in FPH is substantially more expensive than that of DPH. 
The key reason is that FPH does not compute PH on the original graph structure but rather on a \emph{subgraph-pattern-induced filtration} constructed from all frequent patterns identified during FSM.  
DPH relies solely on a degree-based filtration, which operates directly on the graph and produces only $0$- and $1$-dimensional topological features. 
Although significantly faster, DPH cannot capture higher-dimensional topological structures such as $H_2$, which FPH is capable of representing due to its richer pattern-based filtration. 
This highlights a fundamental trade-off between computational efficiency and expressive topological power.
The DD dataset exhibits much larger runtimes than AIDS and PROTEINS across all
FPH-based methods. This is mainly due to its larger graph density, since the complexity of FSM grows quickly with the number of subgraphs and their embeddings.
In addition, the resulting pattern-induced filtrations are much larger, leading to a more expensive PH computation. 

\begin{figure}[ht]
    \centering
    \includegraphics[width=0.65\textwidth]{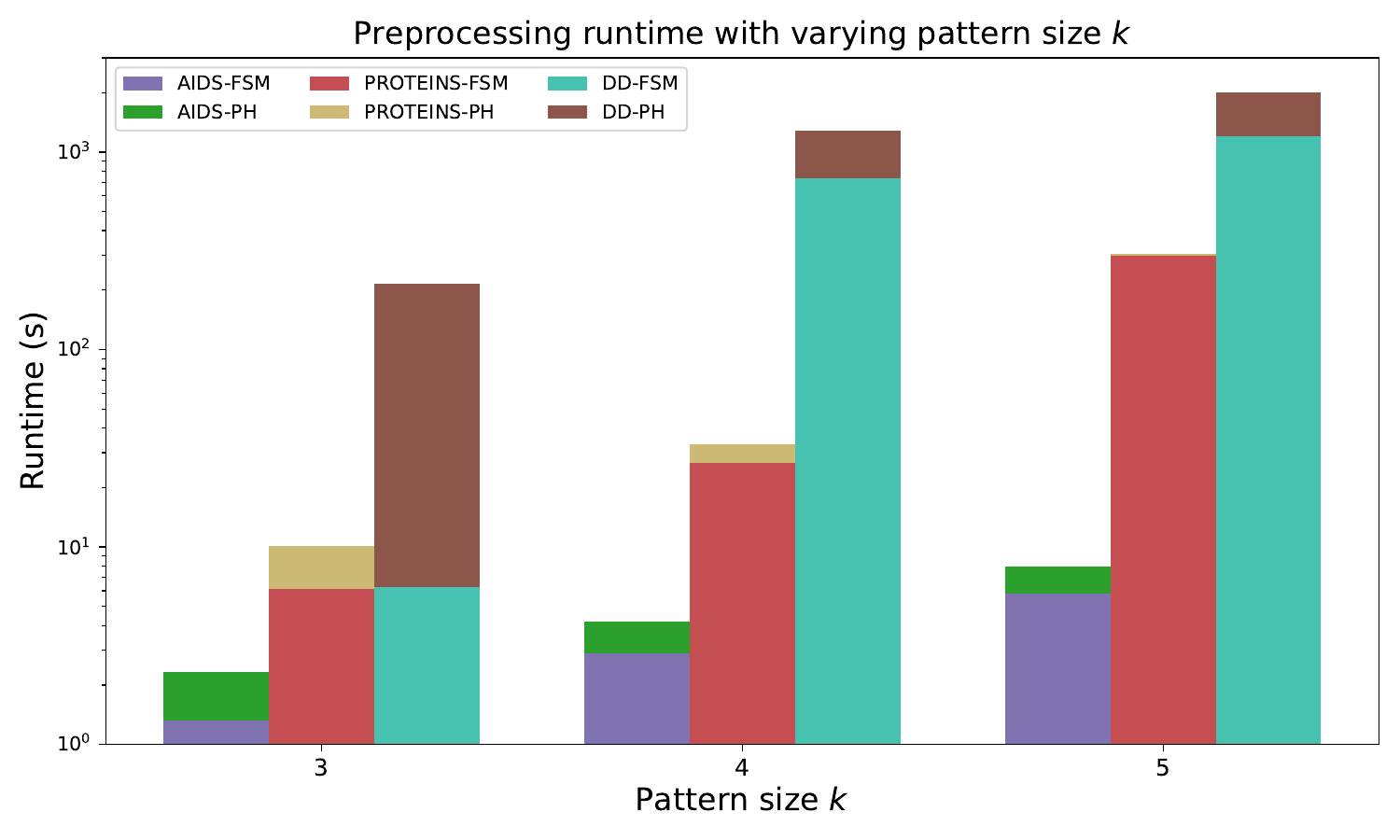} 
    \caption{Preprocessing runtime with varying $k$.}
    \label{fig:runtime_k} 
\end{figure}

Based on the runtime analysis above, the major computational overhead lies in the preprocessing stage. We further evaluate the sensitivity of the method with respect to the pattern size $k$ used in FSM and the subsequent PH construction. To this end, we conduct experiments by varying $k$ from $3$ to $5$.
Figure~\ref{fig:runtime_k} shows that the preprocessing cost increases as the pattern size~$k$ grows. 
Although larger patterns naturally incur higher computational overhead, the cost remains manageable for typical choices of $k$, and small datasets such as AIDS and PROTEINS remain efficient even at $k=5$. 
Importantly, our earlier results indicate that most of the discriminative power already emerges at the 2-dimensional level (i.e., features from $H_0$ to $H_2$). 
Increasing $k$ beyond this range significantly expands the mined pattern space and the induced simplicial complexes, but in many cases does not lead to any noticeable performance improvement.
Therefore, meaningful topological information is captured at relatively small pattern sizes, which keeps the overall preprocessing cost under control.

While the above results demonstrate the effectiveness and scalability of FSF, its practical use on large graph databases also requires controlling the cost
of the preprocessing stage.
We therefore consider a budget-controlled mining strategy in the following section.

\subsection{Budget-controlled approximate FSM}

Exact frequent subgraph mining provides a complete set of patterns, but its computational cost can become expensive in both runtime and memory, particularly on medium- and large-scale graph datasets.
This issue is further exacerbated for node-labeled graphs with low label entropy, where a large number of subgraph embeddings may be generated during pattern growth.

\begin{figure}[htbp]
    \centering
    \includegraphics[width=0.45\textwidth]{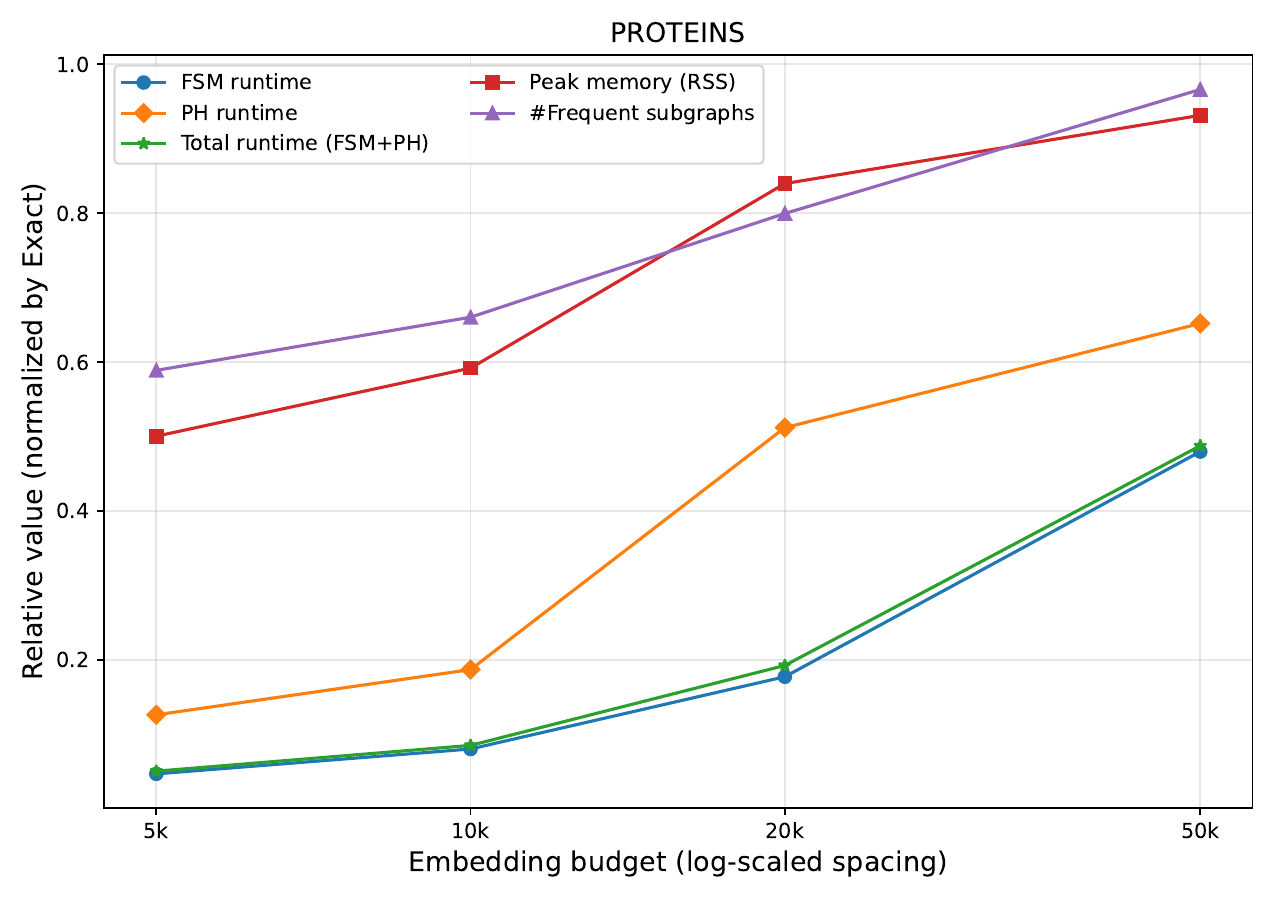} \ \ 
    \includegraphics[width=0.45\textwidth]{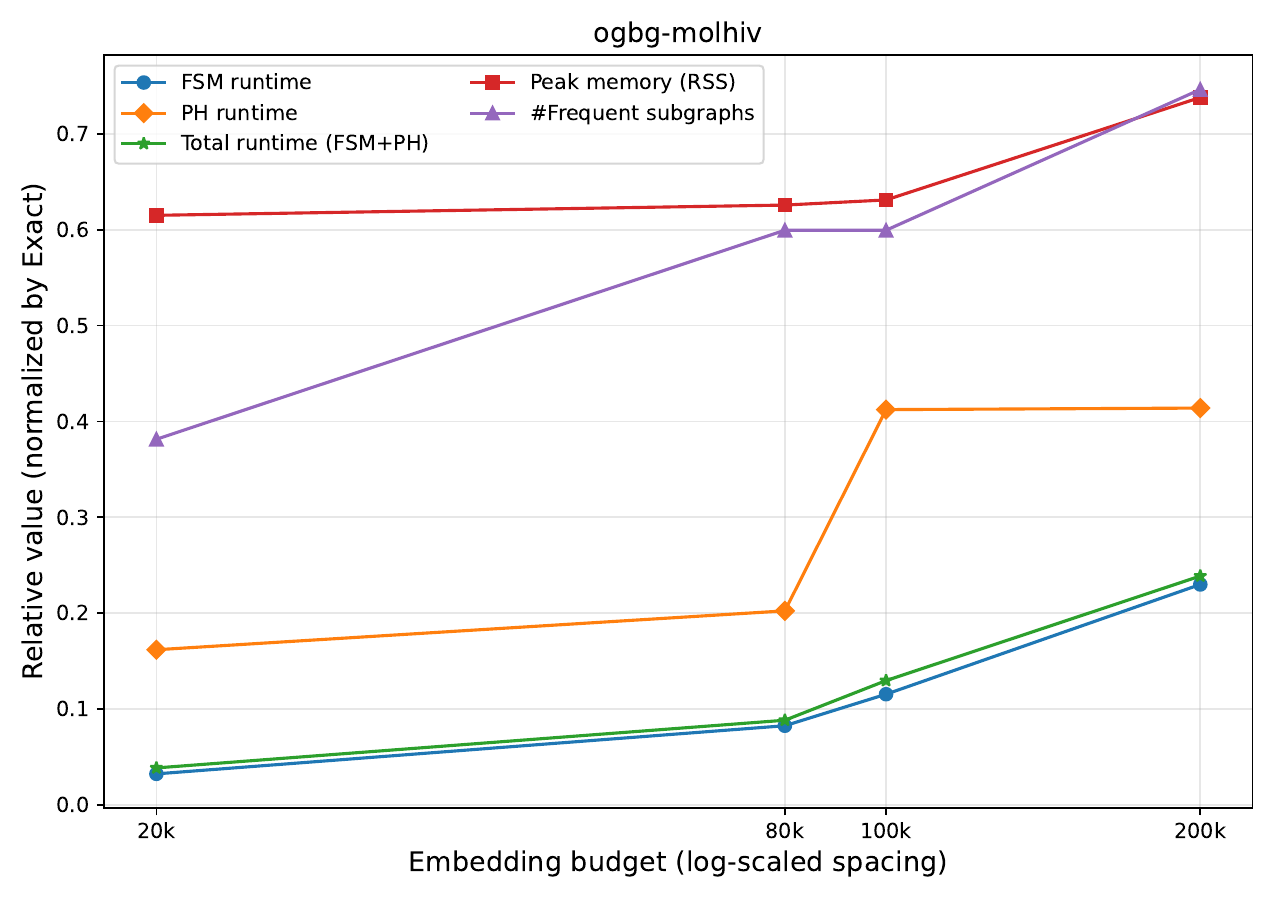} 
   \caption{Effect of embedding budget on frequent subgraph mining and FPH calculation for PROTEINS (left) and ogbg-molhiv (right).} \label{fig:effect_budget} 
\end{figure}

Motivated by these scalability challenges, we investigate a budget-controlled variant of frequent subgraph mining that explicitly limits the number of retained subgraph embeddings for each pattern-growth process.
We evaluate this strategy on two datasets with distinct characteristics:
PROTEINS, a medium-scale dataset with low node-label entropy, and ogbg-molhiv, a large-scale graph dataset with a substantially larger number of graphs.
Figure~\ref{fig:effect_budget} reports the preprocessing runtime, peak resident set size (RSS), and the number of discovered frequent subgraphs under different embedding budgets, normalized by the exact mining result with pattern size $k=5$.
The absolute cost of exact mining is summarized in Table~\ref{tab:exact_cost}.

\begin{table}[htbp]
\centering
\small
\caption{Cost of exact preprocessing.}
\label{tab:exact_cost}
\begin{tabular}{lccccc}
\toprule
Dataset & FSM Runtime (s) & PH Runtime (s) & Total Runtime (s) & Peak RSS (GB) & \#Frequent \\
\midrule
PROTEINS     & 298.0 & 14.2 & 312.2 & 1.31 & 883 \\
ogbg-molhiv  & 617.6 & 31.1   &  648.7  & 1.87 & 674 \\
\bottomrule
\end{tabular}
\end{table}

According to the results shown in Figure~\ref{fig:effect_budget}, as the embedding budget increases, the preprocessing runtime, peak memory consumption, and the number of discovered frequent subgraphs all grow steadily, approaching the cost of exact mining.
This trend is consistent across both datasets, indicating that the budget parameter effectively controls the computational complexity of the mining process.
These results demonstrate that budget-controlled FSM provides a flexible mechanism to balance computational efficiency and information coverage.

To further assess the impact of embedding budget on downstream performance, we evaluate FPH-ML on the PROTEINS dataset under different budget settings.
Since FPH-ML relies solely on FPH features without additional representation learning, it provides a direct and sensitive perspective for examining the effect of budget control.
Table~\ref{tab:proteins_budget_accuracy} reports the corresponding classification results.
Notably, restricting the number of retained subgraph embeddings does not necessarily degrade downstream performance.
In particular, budgets of 5k and 20k (highlighted in bold) achieve accuracy comparable to that of exact mining, indicating that a relatively small subset of subgraphs is sufficient to capture the discriminative topological information required for classification.
Moreover, the non-monotonic performance trend suggests that including additional subgraphs may introduce redundant or weakly discriminative patterns, highlighting the regularization effect of budget-controlled mining.

\begin{table}[htbp]
\centering
\caption{Graph classification accuracy (\%) of FPH-ML on the PROTEINS dataset under different embedding budgets. Results are reported as mean $\pm$ standard deviation over 10 folds.}
\label{tab:proteins_budget_accuracy}
\begin{tabular}{l c}
\toprule
Embedding Budget & Accuracy (\%) \\
\midrule
Exact & $74.31 \pm 4.78$ \\
50k   & $72.96 \pm 3.32$ \\
20k   & $\mathbf{73.86 \pm 2.92}$ \\
10k   & $71.08 \pm 3.49$ \\
5k    & $\mathbf{73.86 \pm 2.39}$ \\
\bottomrule
\end{tabular}
\end{table}

\section{Limitations and future directions}\label{Limitations}

Despite these strengths, the proposed method still has several limitations. In particular, it inherits computational challenges from both frequent subgraph mining (FSM) and persistent homology (PH), and it is not directly applicable to dynamic or evolving graph datasets. We highlight several promising directions to address these challenges.

First, beyond efficiency-oriented approximations, future work may explore \textit{topology-preserving approximate FSM} methods that prioritize subgraphs not only by frequency, but also by their contribution to preserving persistent homological features of the original graphs.
Second, applying \textit{graph coarsening} prior to FSM, while guaranteeing topological similarity to the original graph, may reduce both the search space and the size of the induced filtration.
Third, incorporating \textit{incremental or dynamic FSM} would enable efficient updates when new graphs arrive, making the framework suitable for streaming or evolving datasets without requiring re-mining from scratch.
Finally, future work may explore \textit{end-to-end differentiable variants} of FPH by integrating learnable filtrations or neural estimators of pattern importance, enabling joint optimization with downstream tasks and potentially improving both efficiency and performance.

From a theoretical perspective, we plan to further study the stability properties of frequent subgraph-based filtrations.
In addition, we aim to better understand the topological interplay between frequent patterns and their induced PH features.

\section{Conclusions} \label{Conclusion}

We propose a novel filtration, FSF, for computing persistent homology on graphs. 
It is the first filtration constructed from frequent subgraph patterns mined across the entire dataset, and we also provide the theoretical analysis of its properties. 
In particular, we show that persistence obtained from FSF reflects both the frequency of patterns and the topology of the graph.
FSF enriches the set of PH filtrations and bridges FSM with TDA for graph classification.
Beyond persistent homology itself, we propose two approaches, FPH-ML and FPH-GNN, to incorporate the proposed PH features into graph classification. 
Extensive experiments on multiple benchmark datasets show that our methods outperform kernel-based, GNN-based, and PH-based baselines in most cases, highlighting the benefit of injecting high-order topological features into graph classification models.  



\bibliographystyle{cas-model2-names}
\bibliography{CX-References}

@inproceedings{gilmer2017neural,
  title={Neural message passing for quantum chemistry},
  author={Gilmer, Justin and Schoenholz, Samuel S and Riley, Patrick F and Vinyals, Oriol and Dahl, George E},
  booktitle = {Proc. Intern. Conf. on Machine Learning},
  pages={1263--1272},
  year={2017},
}

@inproceedings{AslayNMG18,
  author       = {{\c{C}}igdem Aslay and
                  Muhammad Anis Uddin Nasir and
                  Gianmarco De Francisci Morales and
                  Aristides Gionis},
  title        = {Mining Frequent Patterns in Evolving Graphs},
  booktitle    = {Proc. of the 27th {ACM} Intern. Conf. on Information
                  and Knowledge Management, {CIKM}},
  pages        = {923--932},
  year         = {2018},
}

@inproceedings{ChenCCG23,
  author       = {Zhaoming Chen and
                  Xinyang Chen and
                  Guoting Chen and
                  Wensheng Gan},
  title        = {Frequent Subgraph Mining in Dynamic Databases},
  booktitle    = {{IEEE} Intern. Conf. on Big Data, BigData},
  pages        = {5733--5742},
  year         = {2023},
}

@article{Chen2024closed,
	author = {Chen, Xinyang and Cai, Jiayu and Chen, Guoting and Gan, Wensheng and Broustet, Amaël},
	year = {2024},
	month = {02},
	pages = {120363},
	title = {{FCSG-Miner}: Frequent closed subgraph mining in multi-graphs},
	volume = {665},
	journal = {Information Sciences},
}

@article{Elseidy2014grami,
	author    = {Mohammed Elseidy and
	Ehab Abdelhamid and
	Spiros Skiadopoulos and
	Panos Kalnis},
	title     = {{GRAMI:} Frequent Subgraph and Pattern Mining in a Single Large Graph},
	journal   = {Proc. VLDB Endowment},
	volume    = {7},
	number    = {7},
	pages     = {517--528},
	year      = {2014},
}

@article{He2024,
author = {Cheng He and Xinyang Chen and Guoting Chen and Wensheng Gan and Philippe Fournier-Viger},
title = {Mining credible attribute rules in dynamic attributed graphs},
journal = {Expert Syst. with Appl.},
volume       = {246}, 
pages = {123012:1-11},
year = {2024},
}

@article{HeCCG25,
  author       = {Cheng He and
                  Jiayu Cai and
                  Guoting Chen and
                  Wensheng Gan},
  title        = {A generic framework for mining sequences with various interestingness
                  measures in dynamic attributed graphs},
  journal      = {Knowl. Inf. Syst.},
  volume       = {67},
  number       = {8},
  pages        = {6689--6716},
  year         = {2025},
}

@inproceedings{inokuchi2000apriori,
  title={An apriori-based algorithm for mining frequent substructures from graph data},
  author={Inokuchi, Akihiro and Washio, Takashi and Motoda, Hiroshi},
  booktitle={Proc. European Conference on Principles of Data Mining and Knowledge Discovery},
  pages={13--23},
  year={2000},
}

@article{Morris2020TUDataset,
  author       = {Christopher Morris and
                  Nils M. Kriege and
                  Franka Bause and
                  Kristian Kersting and
                  Petra Mutzel and
                  Marion Neumann},
  title        = {TUDataset: {A} collection of benchmark datasets for learning with graphs},
  journal     = {arXiv preprint},
  volume       = {2007.08663},
  year         = {2020},
}

@inproceedings{Preti2022,
  author       = {Giulia Preti and
                  Gianmarco De Francisci Morales and
                  Francesco Bonchi},
  title        = {{FreSCo:} Mining Frequent Patterns in Simplicial Complexes},
  booktitle    = {Proc. 22nd The Web Conference},
  pages        = {1444--1454},
  year         = {2022}
}

@article{rehman2024study,
  title   ={A Study on Frequent Subgraph Mining Approaches: Challenges and Future Directions},
  author  ={Rehman, Saif Ur and Khalil, Muhammad Ibrahim and Kundi, Mahwish and AlSaedi, Tahani},
  journal ={Research Updates in Mathematics and Computer Science},
  volume  ={4},
  pages   ={33--63},
  year    = {2024}
}

@article{ying2024,
      title={Representation Learning for Frequent Subgraph Mining}, 
      author={Rex Ying and Tianyu Fu and Andrew Wang and Jiaxuan You and Yu Wang and Jure Leskovec},
      year={2024},
      volume ={2402.14367},
      journal={arXiv preprint},
}

@inproceedings{yan2002gSpan,
	author = {Yan, Xifeng and Han, Jiawei},
	year = {2002},
	pages = {721- 724},
	title = {g{S}pan: Graph-Based Substructure Pattern Mining},
	booktitle={Proc. IEEE Intern. Conf. on Data Mining},
}

@INPROCEEDINGS{Shaul2021cgSpan,
	author={Shaul, Zevin and Naaz, Sheikh},
	booktitle={Proc. IEEE Intern. Conf. on Big Data}, 
	title={{cgSpan}: Closed Graph-Based Substructure Pattern Mining}, 
	year={2021},
	pages={4989-4998},
}

@inproceedings{ZeghinaLBV23,
  author       = {Assaad Oussama Zeghina and
                  Aur{\'{e}}lie Leborgne and
                  Florence Le Ber and
                  Antoine Vacavant},
  title        = {{Multi-SPMiner:} {A} Deep Learning Framework for Multi-Graph Frequent
                  Pattern Mining with Application to spatiotemporal Graphs},
  booktitle    = {Proc. Intern. Conf. on Knowledge-Based and Intelligent Information {\&} Engineering Systems (KES)},
  series       = {Procedia Computer Science},
  volume       = {225},
  pages        = {1094--1103},
  year         = {2023},
}

@article{adams2017persistence,
  title={Persistence images: A stable vector representation of persistent homology},
  author={Adams, Henry and Emerson, Tegan and Kirby, Michael and Neville, Rachel and Peterson, Chris and Shipman, Patrick and Chepushtanova, Sofya and Hanson, Eric and Motta, Francis and Ziegelmeier, Lori},
  journal={J. Machine Learning Research},
  volume={18},
  number={8},
  pages={1--35},
  year={2017}
}

@article{DAktasAF19,
  author       = {Mehmet Emin Aktas and
                  Esra Akbas and
                  Ahmed El Fatmaoui},
  title        = {Persistence homology of networks: methods and applications},
  journal      = {Appl. Netw. Sci.},
  volume       = {4},
  number       = {1},
  pages        = {61:1--61:28},
  year         = {2019},
  
}

@article{corcoran2023topological,
  title={Topological data analysis for geographical information science using persistent homology},
  author={Corcoran, Padraig and Jones, Christopher B},
  journal={International J. Geographical Information Science},
  volume={37},
  number={3},
  pages={712--745},
  year={2023},
}

@article{chowdhury2018functorial,
  title={A functorial Dowker theorem and persistent homology of asymmetric networks},
  author={Chowdhury, Samir and M{\'e}moli, Facundo},
  journal={J. Applied and Computational Topology},
  volume={2},
  number={1},
  pages={115--175},
  year={2018},
}

@article{CorcoranJ23,
  author       = {Padraig Corcoran and
                  Christopher B. Jones},
  title        = {Topological data analysis for geographical information science using
                  persistent homology},
  journal      = {Int. J. Geogr. Inf. Sci.},
  volume       = {37},
  number       = {3},
  pages        = {712--745},
  year         = {2023},
}

@article{edelsbrunner2008persistent,
  title={Persistent homology-a survey},
  author={Edelsbrunner, Herbert and Harer, John and others},
  journal={Contemporary mathematics},
  volume={453},
  number={26},
  pages={257--282},
  year={2008},
}

@inproceedings{FugacciSIF16,
  author       = {Ulderico Fugacci and
                  Sara Scaramuccia and
                  Federico Iuricich and
                  Leila De Floriani},
  title        = {Persistent Homology: a Step-by-step Introduction for Newcomers},
  booktitle    = {Proc. Conf. Smart Tools and Apps in Computer Graphics, {STAG}},
  pages        = {1--10},
  year         = {2016},
}

@article{flammer2024spatiotemporal,
  title={Spatiotemporal Persistence Landscapes},
  author={Flammer, Martina and H{\"u}per, Knut},
  journal={arXiv preprint},
  volume={2412.11925},
  year={2024}
}

@article{immonen2023going,
  title={Going beyond persistent homology using persistent homology},
  author={Immonen, Johanna and Souza, Amauri and Garg, Vikas},
  journal={Advances in neural information processing systems},
  volume={36},
  pages={63150--63173},
  year={2023}
}

@article{liang2021analysis,
  title={Analysis of brain functional connectivity neural circuits in children with autism based on persistent homology},
  author={Liang, Di and Xia, Shengxiang and Zhang, Xianfu and Zhang, Weiwei},
  journal={Frontiers in Human Neuroscience},
  volume={15},
  pages={745671},
  year={2021},
}

@article{petri2013topological,
  title={Topological strata of weighted complex networks},
  author={Petri, Giovanni and Scolamiero, Martina and Donato, Irene and Vaccarino, Francesco},
  journal={PloS one},
  volume={8},
  number={6},
  pages={e66506},
  year={2013},
}

@article{pun2018persistent,
  title={Persistent-Homology-based machine learning and its applications--A survey},
  author={Pun, Chi Seng and Xia, Kelin and Lee, Si Xian},
  journal={arXiv preprint},
  volume = {1811.00252},
  year={2018}
}

@article{pham2025topological,
  title={Topological data analysis in graph neural networks: Surveys and perspectives},
  author={Pham, Phu and Bui, Quang-Thinh and Nguyen, Ngoc Thanh and Kozma, Robert and Yu, Philip S and Vo, Bay},
  journal={IEEE Trans. Neural Networks and Learning Systems},
  volume       = {36},
  number       = {6},
  pages        = {9758--9776},
  year={2025},
}

@article{ravishanker2021introduction,
  title={An introduction to persistent homology for time series},
  author={Ravishanker, Nalini and Chen, Renjie},
  journal={Wiley Interdisciplinary Reviews: Computational Statistics},
  volume={13},
  number={3},
  pages={e1548},
  year={2021},
}

@article{wadhwa2018flat,
  title={A flat persistence diagram for improved visualization of persistent homology},
  author={Wadhwa, Raoul R and Dhawan, Andrew and Williamson, Drew FK and Scott, Jacob G},
  journal={arXiv preprint},
  volume = {1812.04567},
  year={2018}
}

@inproceedings{zhao2020persistence,
  title={Persistence enhanced graph neural network},
  author={Zhao, Qi and Ye, Ze and Chen, Chao and Wang, Yusu},
  booktitle={Proc. Intern. Conf. on Artificial Intelligence and Statistics},
  pages={2896--2906},
  year={2020},
}

@article{hamilton2017inductive,
  title={Inductive representation learning on large graphs},
  author={Hamilton, Will and Ying, Zhitao and Leskovec, Jure},
  journal={Advances in neural information processing systems},
  volume={30},
  year={2017}
}

@incollection{kashima2004kernels,
  title={Kernels for graphs},
  author={Kashima, Hisashi and Tsuda, Koji and Inokuchi, Akihiro},
  booktitle={Kernel methods in computational biology},
  pages={155--170},
  year={2004},
}

@article{KipfW16,
  author       = {Thomas N. Kipf and Max Welling},
  title        = {Semi-Supervised Classification with Graph Convolutional Networks},
  journal      = {arXiv preprint},
  volume       = {1609.02907},
  year         = {2016},
}

@article{kriege2020survey,
  title={A survey on graph kernels},
  author={Kriege, Nils M and Johansson, Fredrik D and Morris, Christopher},
  journal={Applied Network Science},
  volume={5},
  number={1},
  pages={6},
  year={2020},
}

@article{narayanan2017graph2vec,
  title={graph2vec: Learning distributed representations of graphs},
  author={Narayanan, Annamalai and Chandramohan, Mahinthan and Venkatesan, Rajasekar and Chen, Lihui and Liu, Yang and Jaiswal, Shantanu},
  journal={arXiv preprint},
  volume={1707.05005},
  year={2017}
}

@inproceedings{shervashidze2009efficient,
  title={Efficient graphlet kernels for large graph comparison},
  author={Shervashidze, Nino and Vishwanathan, SVN and Petri, Tobias and Mehlhorn, Kurt and Borgwardt, Karsten},
  booktitle={Artificial intelligence and statistics},
  pages={488--495},
  year={2009},
}

@inproceedings{ju2025cluster,
  title={Cluster-guided contrastive class-imbalanced graph Classification},
  author={Ju, Wei and Mao, Zhengyang and Yi, Siyu and Qin, Yifang and Gu, Yiyang and Xiao, Zhiping and Shen, Jianhao and Qiao, Ziyue and Zhang, Ming},
  booktitle={Proc. AAAI Conference on Artificial Intelligence},
  volume={39},
  pages={11924--11932},
  year={2025}
}

@article{liu2023survey,
  title={A survey on graph classification and link prediction based on {GNN}},
  author={Liu, Xingyu and Chen, Juan and Wen, Quan},
  journal={arXiv preprint},
  volume = {2307.00865},
  year={2023}
}

@article{shervashidze2011weisfeiler,
  title={Weisfeiler-lehman graph kernels.},
  author={Shervashidze, Nino and Schweitzer, Pascal and Van Leeuwen, Erik Jan and Mehlhorn, Kurt and Borgwardt, Karsten M},
  journal={J. Machine Learning Research},
  volume={12},
  number={9},
  year={2011}
}

@inproceedings{shi2025two,
  title={Two-View Fusion Graph Neural Networks for Graph Classification},
  author={Shi, Zhouhua and Sun, Shiwen and Yang, Guang and Liu, Yan},
  booktitle={Proc. Intern. Conf. on Intelligent Computing},
  pages={185--196},
  year={2025},
}

@article{velivckovic2017graph,
  title={Graph attention networks},
  author={Veli{\v{c}}kovi{\'c}, Petar and Cucurull, Guillem and Casanova, Arantxa and Romero, Adriana and Lio, Pietro and Bengio, Yoshua},
  journal={arXiv preprint},
  volume ={1710.10903},
  year={2017}
}

@inproceedings{XuHLJ19,
  author       = {Keyulu Xu and
                  Weihua Hu and
                  Jure Leskovec and
                  Stefanie Jegelka},
  title        = {How Powerful are Graph Neural Networks?},
  booktitle    = {Proc. 7th Intern. Conf. on Learning Representations},
  pages   = {OpenReview.net},
  year         = {2019},
}

@article{hu2020open,
  title={Open graph benchmark: Datasets for machine learning on graphs},
  author={Hu, Weihua and Fey, Matthias and Zitnik, Marinka and Dong, Yuxiao and Ren, Hongyu and Liu, Bowen and Catasta, Michele and Leskovec, Jure},
  journal={Advances in neural information processing systems},
  volume={33},
  pages={22118--22133},
  year={2020}
}

@article{horn2021topological,
  title={Topological graph neural networks},
  author={Horn, Max and De Brouwer, Edward and Moor, Michael and Moreau, Yves and Rieck, Bastian and Borgwardt, Karsten},
  journal={arXiv preprint arXiv:2102.07835},
  year={2021}
}

@article{yan2025enhancing,
  title={Enhancing graph representation learning with localized topological features},
  author={Yan, Zuoyu and Zhao, Qi and Ye, Ze and Ma, Tengfei and Gao, Liangcai and Tang, Zhi and Wang, Yusu and Chen, Chao},
  journal={J. Machine Learning Research},
  volume={26},
  number={5},
  pages={1--36},
  year={2025}
}

\end{document}